\documentclass[lettersize,journal]{IEEEtran}
\usepackage{amsmath,amsfonts}
\usepackage{algorithmic}
\usepackage{algorithm}
\usepackage{array}
\usepackage[caption=false,font=normalsize,labelfont=sf,textfont=sf]{subfig}
\usepackage{textcomp}
\usepackage{stfloats}
\usepackage{url}
\usepackage{verbatim}
\usepackage{graphicx}
\usepackage{cite}
\usepackage{multirow}
\usepackage{xcolor}

\definecolor{forestgreen}{RGB}{34,139,34}
\begin{document}

\title{A Cluster-Based Trip Prediction Graph Neural Network Model for Bike Sharing Systems}

\author{Bárbara Tavares, Cláudia Soares, Manuel Marques~\IEEEmembership{}
    % <-this % stops a space

%\thanks{This paper was produced by the IEEE Publication Technology Group. They are in Piscataway, NJ.}% <-this % stops a space
%\thanks{Manuscript received April 19, 2021; revised August 16, 2021.}}

\thanks{Bárbara Tavares and Manuel Marques are with Instituto Superior Técnico and Cláudia Soares is with NOVA School of Science and Technology. 

This work has been submitted to the IEEE for possible publication. Copyright may be transferred without notice, after which this version may no longer be accessible.}}



\maketitle

\begin{abstract}
Bike Sharing Systems (BSSs) are emerging as an innovative transportation service. Ensuring the proper functioning of a BSS is crucial given that these systems are committed to eradicating many of the current global concerns, by promoting environmental and economic sustainability and contributing to improving the life quality of the population.
Good knowledge of users' transition patterns is a decisive contribution to the quality and operability of the service.
The analogous and unbalanced users’ transition patterns cause these systems to suffer from bicycle imbalance, leading to a drastic customer loss in the long term. Strategies for bicycle rebalancing become important to tackle this problem and for this, bicycle traffic prediction is essential, as it allows to operate more efficiently and to react in advance.
In this work, we propose a bicycle trips predictor based on Graph Neural Network embeddings, taking into consideration station groupings, meteorology conditions, geographical distances, and trip patterns. We evaluated our approach in the New York City BSS ({\it{CitiBike}}) data and compared it with four baselines, including the non-clustered approach. To address our problem's specificities, we developed the Adaptive Transition Constraint Clustering Plus (AdaTC\(_+\)) algorithm, eliminating shortcomings of previous work. Our experiments evidence the clustering pertinence (88\% accuracy compared with 83\% without clustering) and which clustering technique best suits this problem. Accuracy on the Link Prediction task is always higher for AdaTC\(_+\) than benchmark clustering methods when the stations are the same, while not degrading performance when the network is upgraded, in a mismatch with the trained model.
\end{abstract}


\section{Introduction}
\IEEEPARstart{B}{ike} Sharing Systems are prominent, especially in major cities across Europe, Asia, and America, as evidenced by the Meddin Bike-Sharing World Map \cite{worldmap:bs}. At the time of writing, this map accounted for almost 10 million bikes worldwide and within the active BSS, about 65.53\% maintain the activity, 33.07\% no longer operate, and the remaining 1.4\% relate to temporarily or seasonally closed stations \cite{DeMaio:blog, DeMaio2009}. A BSS is a service that offers an alternative form of mobility, in which bicycles are made available for shared use on particular trips, which can start and end at any geographical point, as long as there are bicycles on the departure local/station and valid empty places on the return local/dock. The two systems just described are a Dockless Bike Sharing System (DBSS) and a Station Bike Sharing System (SBSS), respectively. 

\begin{figure}[!t]
\centering
\subfloat[]{\includegraphics[width=1.7in,height=1.3in]{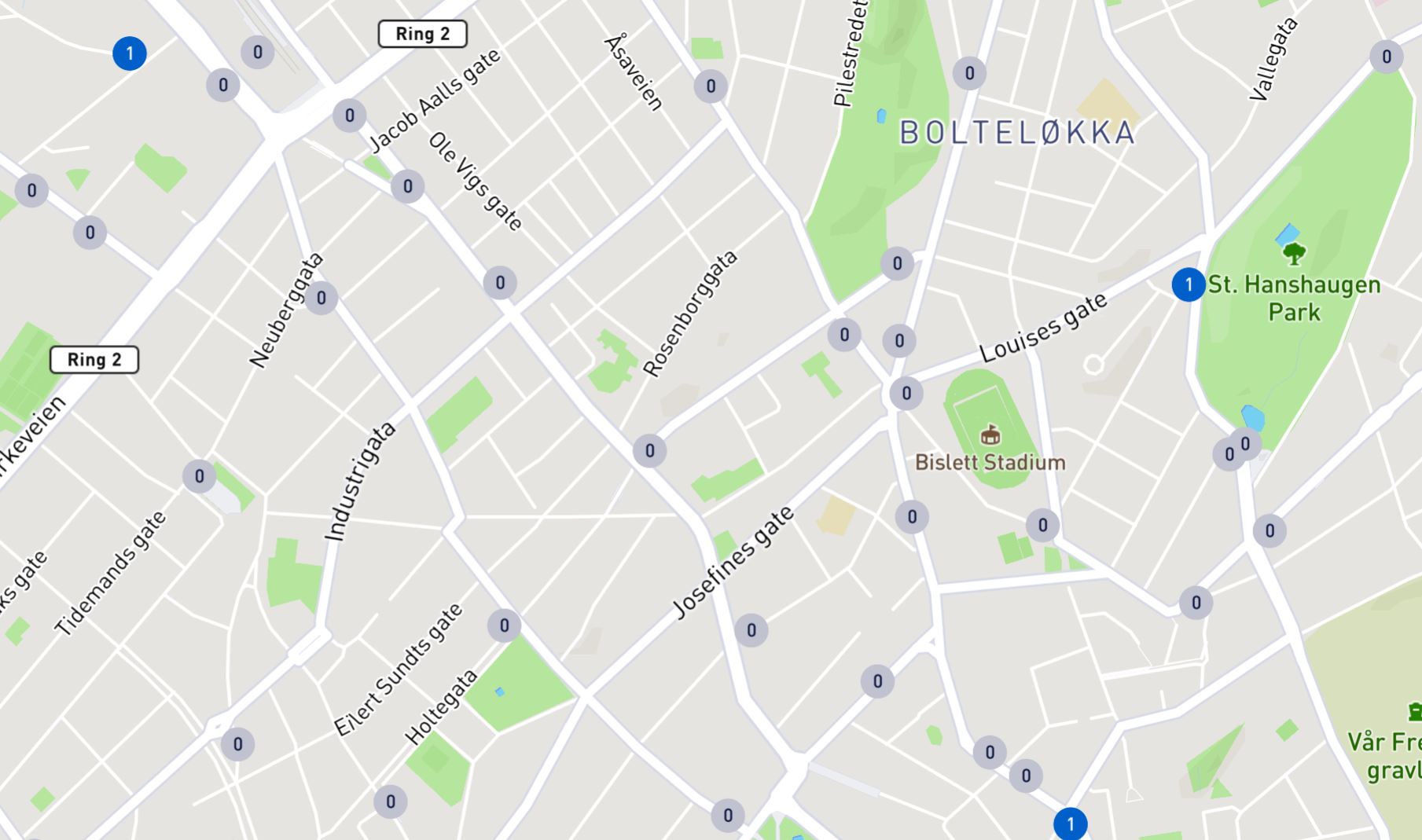}%
\label{fig:fig151}}
\hfil
\subfloat[]{\includegraphics[width=1.7in,height=1.3in]{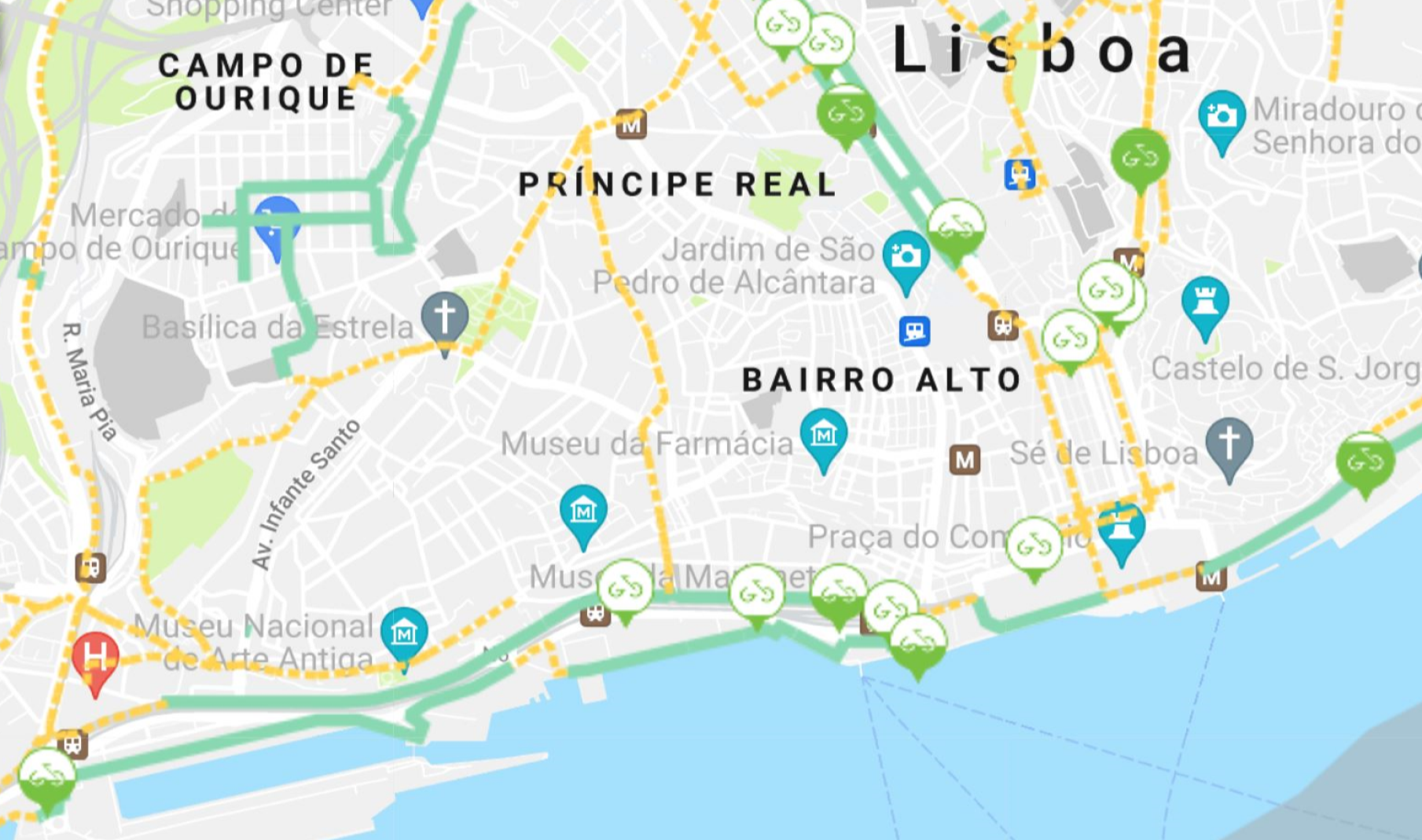}%
\label{fig:fig152}}
\caption{Absence or Reduced Supply of Bicycles  for Rent in a Wide Area in the SBSSs (a) {\it{Oslo Bysykkel}} in Oslo and (b) {\it{GIRA}} in Lisbon.}
\label{fig:fig15}
\end{figure}

Both SBSS and DBSS suffer from a common problem: the bike imbalance due to analogous users' transition patterns, leaving not only empty stations without bikes available to rent but also congested stations without available docks to park \cite{Beecham2014}. This is evidenced in Figure \ref{fig:fig15} for two different SBSS, where for each one we present a live count of the bikes available to rent in some stations. There are some stations at the full occupation capacity and a wide area with stations without bikes available to rent and for those that have, the supply is residual. 

This problem leads to a drastic customer loss in the long term, and for DBSS may cause additional issues related to the service level \cite{Fishman2015,Brien2014},  {\it{e.g.}}, inappropriate parking in pedestrian sidewalks and roads, causing significant urban clutter. 
Facing this scenario, in the {\bf{System Operation}} branch, it is important to develop efficient strategies of bike rebalancing, maintaining good operability and service quality of these systems. The {\bf{System Prediction}} branch is entirely linked to the previous one: an accurate model for predicting the traffic should be part of the solution of a good rebalancing strategy, as it allows to operate a BSS more efficiently, improving resource utilization, and allowing to react in advance, decreasing the uncertainty {\it{e.g.}}, in the reallocation process. 
Therefore, we address the Bike Traffic Prediction Problem and our contributions can be summarized as follows:
\begin{itemize}
    \item{An Adaptive Transition Constraint Clustering Plus ({\bf{AdaTC\(_+\)}}) algorithm is proposed to generate clusters of stations in a SBSS, that are grouped considering their geographical distance and transition patterns. {\bf{AdaTC\(_+\)}} extends the previous work \cite{Li2020} by addressing the shortcomings mentioned in Section \ref{sec:paralellism}. }
    \item{A bicycle trips binary predictor (Link Prediction Model) to predict trips, based on Graph Neural Network (GNN) embeddings, taking into consideration the stations groups according to {\bf{AdaTC\(_+\)}} and other clustering techniques, the meteorology conditions, and trip patterns, namely the relative frequency histogram of trips during the days of the week and in predefined time periods.}
    \item{We evaluate our model on a dataset from the NYC BSS, {\it{CitiBike}}, in a test set for 2018 and 2019. We also evaluate the clustering pertinence and compare the results of our predictor model with the results obtained by integrating the groups formed according to three other clustering baselines and a baseline without clustering. The baseline that best suits this data and problem is determined by evaluating the accuracy of the respective Link Prediction (LP) Model.}
\end{itemize}
Our method combines an accurate GNN with a robust clustering approach on different trip features. By following this approach we add robustness to an accurate prediction \cite{8438463,Jia2019,Li2020}. In addition, it can be generalizable and extendable for a wide spectrum of mobility system, depending on the type of data that the system provides.

\subsection{Challenges when Predicting Bike Traffic}
\begin{figure}[!t]
\centering
\includegraphics[width=3.4in]{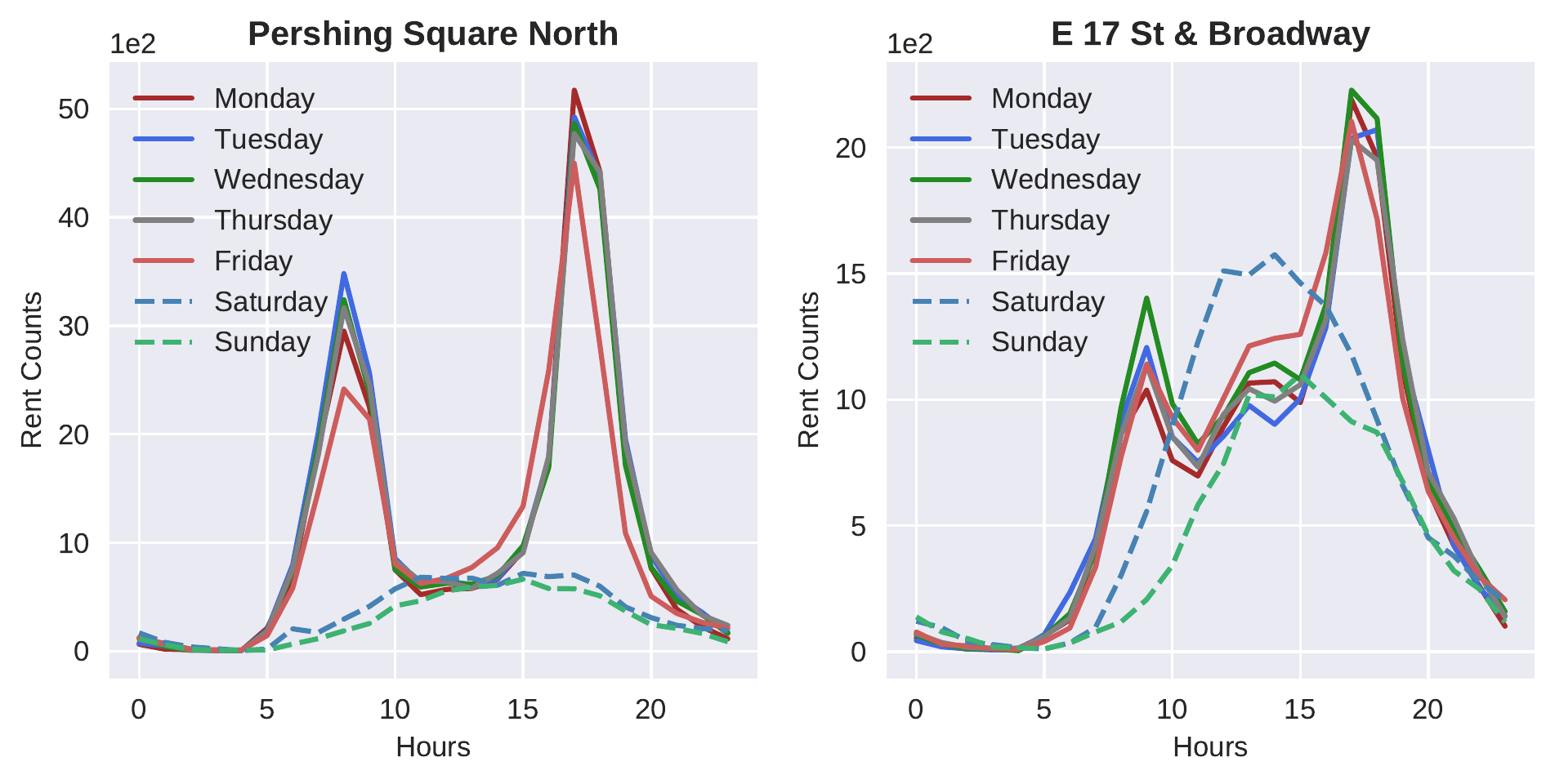}
\caption{Hourly and Weekly Bike Demand at two {\it{CitiBike}} Stations.}
\label{fig:fig1}
\end{figure}

\begin{figure}[!t]
\centering
\includegraphics[width=3.4in]{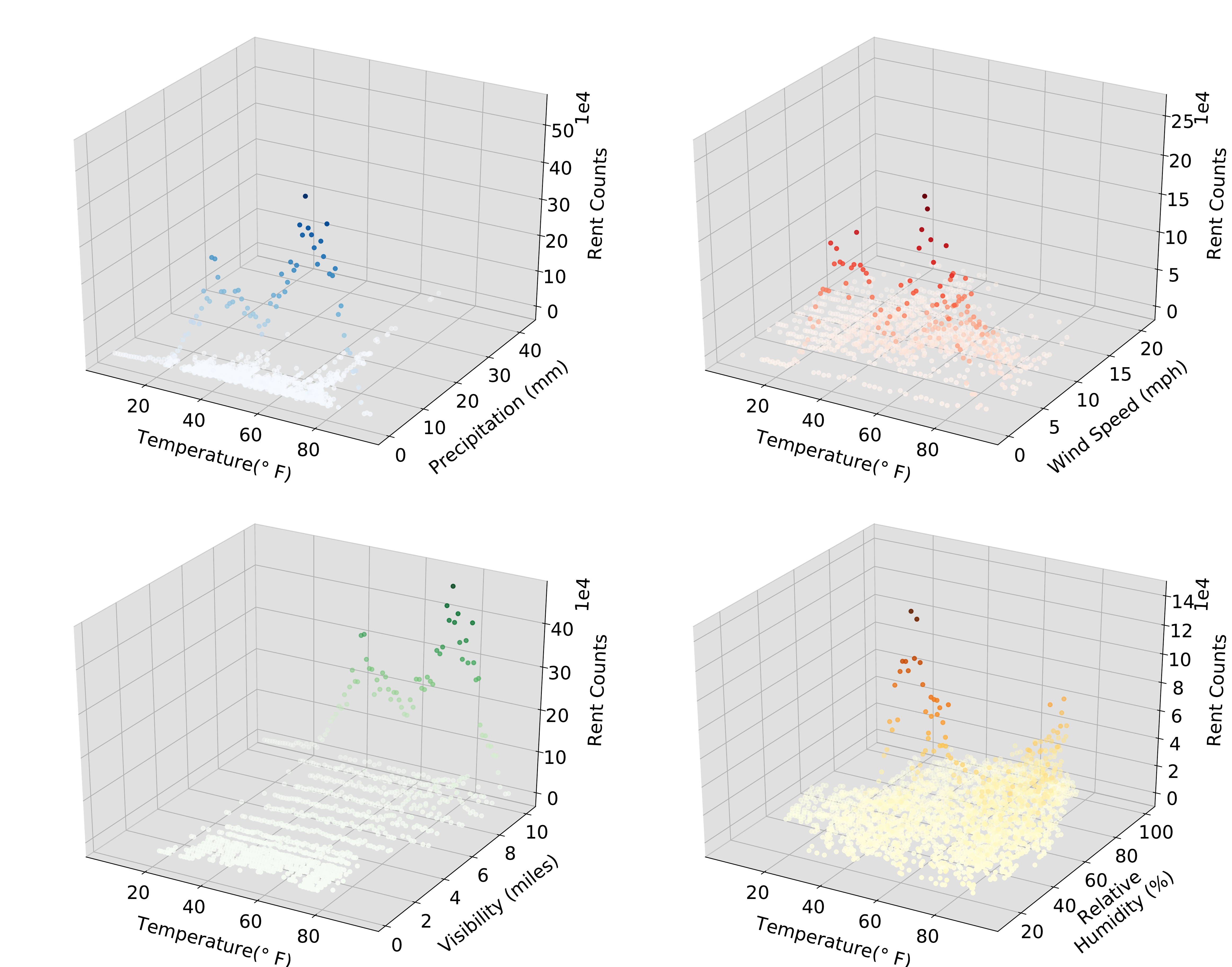}
\caption{Pairwise Impact of Some Weather Conditions on Historical Rents \small{The gradation of colors is according to the z-axis.}}
\label{fig:fig2}
\end{figure}

\begin{figure}[!t]
\centering
\includegraphics[width=3in]{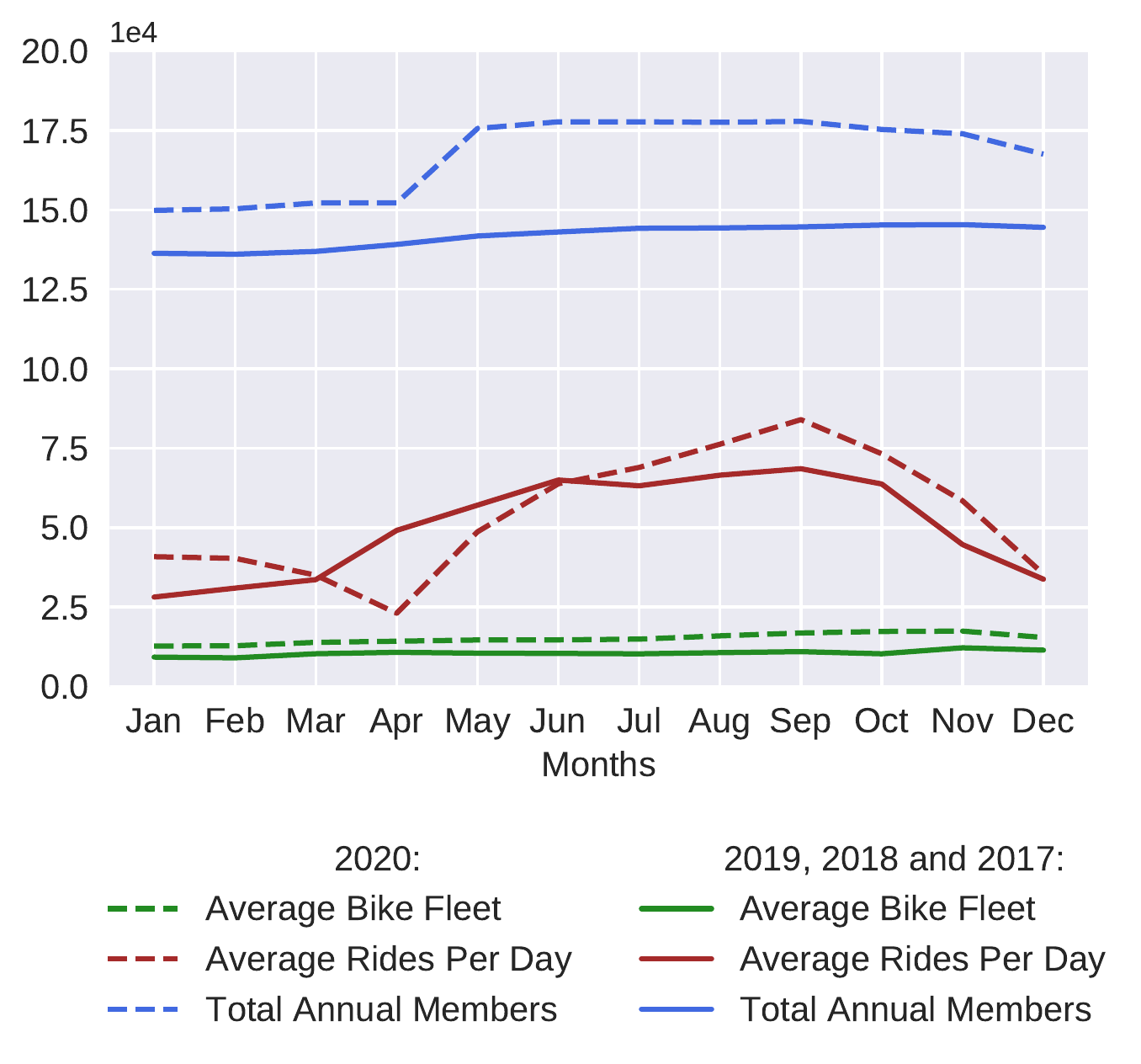}
\caption{Descriptive Measures of the {\it{CitiBike}} service between 2020 and the three years preceding.}
\label{fig:fig3}
\end{figure}

The effectiveness of the bike system rebalancing strategy highly depends on the transitions forecasts accuracy, which faces two main challenges: the demand uncertainty and rides randomness.
The challenges mentioned are illustrated in Figures \ref{fig:fig1}, \ref{fig:fig2} and \ref{fig:fig3}, based on the data used in this work ({\it{CitiBike}} and meteorology data), and will be further detailed in the sequence.

{\bf{Demand Uncertainty}} The users' demand changes temporally and spatially, as can be verified in Figure \ref{fig:fig1}, showing the historical rent fluctuations during different hours and days of the week, for two stations that are 1.95 km away. There is a clear difference in the usage patterns of the two stations during the days of the week (particularly on weekends) and during the day, where the peak hours and activity between high demand peaks differ greatly.
Furthermore, the users' demand is also greatly influenced by external factors such as weather conditions and unexpected events (such as festivals, accidents, the COVID-19 pandemic, etc.) that seldomly occur. Figure~\ref{fig:fig2} represents different weather conditions in which historical trips took place, confirming the inextricable link between weather and the propensity to cycle. In particular, temperature, precipitation, and wind speed seem to be the most determining factors, as expected.

The indicators presented in Figure \ref{fig:fig3} allow to, at a first glance, analyze the impact of the COVID-19 pandemic in {\it{CitiBike}} (officially declared as such by the World Health Organization in March 2020) by comparing the 2020 indicators with the average of these indicators observed in the three years preceding 2020: from 2017 to 2019. Figure \ref{fig:fig3} shows that 2020 began with a decreasing tendency in the average number of rides per day until April, corresponding to the NY government declaration of a full lockdown (from March to April 2020), followed by a four-phase reopening plan from April to July 2020. The reopening may justify the growing trend from April to September when the (global) maximum value of rides per day was reached, surpassing the number of rides in September 2019, 2018, and 2017. The decrease at the end of both years should be mainly justifiable by the winter months, less conducive to the use of this transport (being the average number of rides in 2020 always higher than in the average between 2019, 2018, and 2017).
Beyond the aforementioned facts, in May 2020 the number of annual members increased by almost 17\% when compared to April 2020. {\it{CitiBike}} saw a massive increase of members in pandemic times, probably for being safer transportation in times when social distancing is required. However, it should be noted that the supply of bicycles rose more than 30\% in 2020, compared to the average of 2017, 2018, and 2019 (the plot scale in Figure~\ref{fig:fig3} does not allow to show this growth). Therefore, the significant growth of the system that occurred in 2020 can be justified by the combination of the two factors just mentioned.

{\bf{Rides Randomness}} The vast majority of trips seem to be random if analyzed between individual stations as any user can travel to a nearby station to rent or return a bike. A cluster strategy, that partitions the geographical area under analysis into small functional regions, allows turning infrequent inter-station transitions into frequent intra and inter-cluster transitions, reducing trips' randomness and complexity. 

\section{Related Work}
Among many previous works addressing diversified strategies in other BSSs study branches, such as \textbf{System Pattern Analysis} and \textbf{Operation} \cite{Karlaftis2011,8215489}, we highlight some literature in the \textbf{System Prediction} branch, that focuses on providing forecasts for important indicators, such as predicting the number of bikes, docks available at each station, the entire traffic, etc.

Froehlich et al. \cite{Froehlich2009} compare four predictive models for the bike availability in each station: Bayesian Network, Last Value, Historic Mean, and Historic Trend. Motoaki and Daziano \cite{Motoaki2018} analyze how weather conditions and other factors affect the propensity of cyclists to ride.
Kaltenbrunner et al. \cite{Grivolla} predict the number of bikes and docks available at a given station and time by comparing the models: (1) prediction based on the current state of the station and aggregate statistics of the station’s usage patterns, (2) extrapolating from the current state using the tendencies registered on chosen dates, and (3) an Auto-Regressive Moving Average (ARMA) time series model, that takes into consideration information from surrounding stations.
Many studies consider the relevance of clustering, such as Feng et al. \cite{Feng2018} that conducts station clustering with an iterative spectral clustering, followed by a gradient boosting regression tree to predict the total check-out of the whole BSS. Chen et al. \cite{Chen2016} focused on a dynamic cluster-based framework for over-demand prediction: a weighted correlation network models the relationship among bike stations, grouping neighboring stations with similar bike usage patterns. A Monte Carlo simulation is used to predict the over-demand probability of each cluster. Li et al. \cite{Li2015} modeled the problem as a Hierarchical Prediction Model to forecast the number of bikes that will be rented from and returned to each station, through the flow of bicycles from stations predicted as crowded to stations predicted as empty. This can be considered a Hierarchical Prediction Problem since its structure follows the reasoning of predicting information of the root of the hierarchy (recurring to a Gradient Boosting Regression Tree), predicting the proportion of the distribution at levels below the root (by a Multi-Similarity-Based Inference Model) and use this information to subsequently calculate the number of bicycles. 
The methodology was updated by Li et al.~\cite{Li2020} to predict the city bike usage patterns with a Hierarchical Consistency Prediction Model, formulating the problem as a Hierarchy Time Series Prediction Problem, ensuring that the value at parent nodes is always equal to the sum of its children nodes --- the Hierarchical Consistency property. A three-level hierarchy of locations is formed according to the previous clustering results by the same authors \cite{Li2015}. The rent demand at each level of the hierarchy is separately predicted using a Similarity-based Gaussian Process Regressor.

Machine Learning (ML) and Neural Networks (NN) have been used for road traffic prediction, and a few approaches used clustering \cite{Chen2016,Kim2021,8928512} and Graph Convolutional Neural Networks (GCNN) \cite{09a30cfd5cfe48a49c45c325eec024bc}. Introducing graph information into urban problems is justified by the complex network structure of cities and has been shown to improve prediction results \cite{zhang2017deep,yao2018deep,9112608,Kim2020}. Yang et al. \cite{Yang2020} use an approach fusing measures of network connectivity information as inputs to different ML algorithms to predict short-term bike demand. 
Deep learning-based models have received much attention in the traffic forecasting context and generated state-of-the-art performance. Cui et al. \cite{Cui2020} design a NN structure for traffic forecasting: a stacked bidirectional LSTM, that captures temporal dependencies in Spatio-temporal data, and unidirectional LSTM network architecture (SBU-LSTM). This approach improves traffic prediction accuracy and robustness and presents a data imputation mechanism in the LSTM structure.
Lin et al. \cite{Lin2018} suggest a novel GCNN with a Data-driven Graph Filter that can learn hidden heterogeneous pairwise correlations between stations to predict station-level hourly demand in a BSS.
Zhang et al. \cite{Zhang} came up with a deep Spatio-temporal residual network to collectively predict the inflow and outflow (converted into a 2-channel image-like matrix of crowds) based on historical trajectory data, weather, and events.

\section{Adaptive Transition Constraint Clustering Plus (AdaTC\(_+\))}
\begin{figure*}[!t]
\centering
\includegraphics[width=0.8\linewidth]{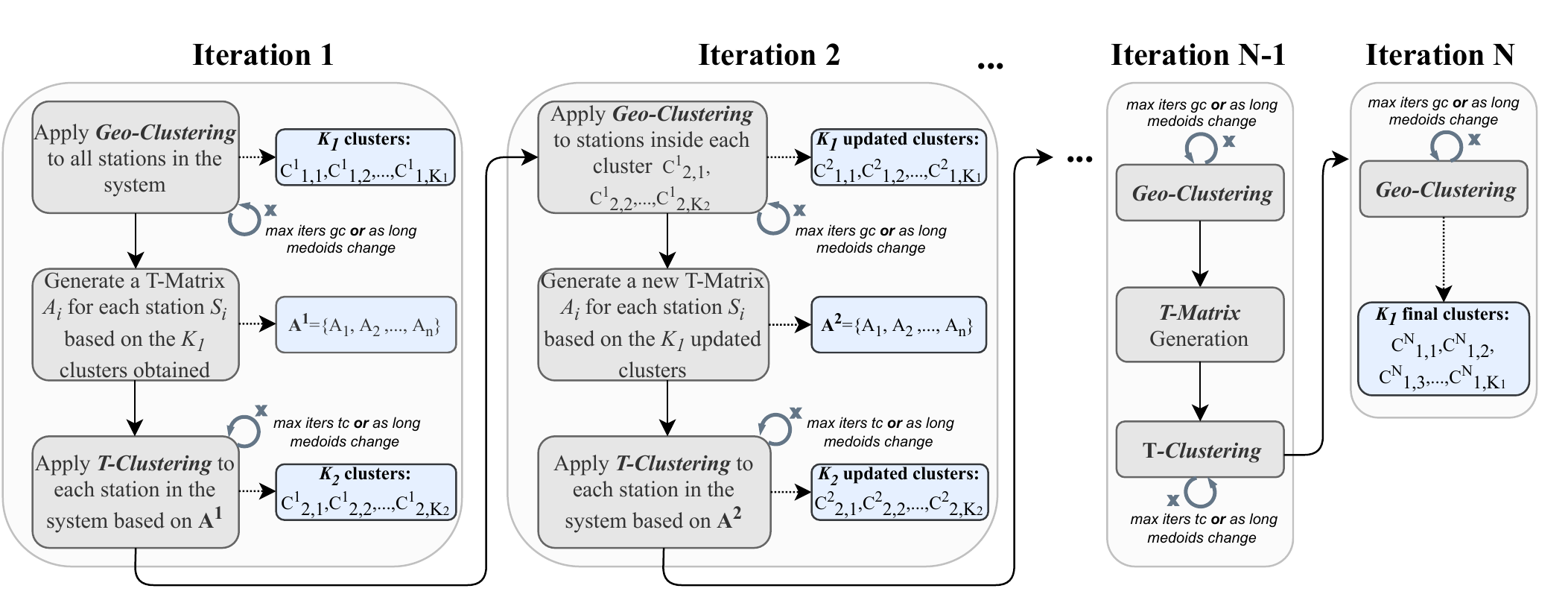}
\caption{{\bf{AdaTC\(_+\)}} Methodology Diagram.}
\label{fig:fig4}
\end{figure*}

The {\bf{AdaTC\(_+\)}} algorithm clusters stations according to their geographical distances and transition patterns, meaning that geographically close stations with similar transition patterns are expected to be grouped. It consists of three main steps: {\it{Geo-Clustering (GC)}}, {\it{Transit-Matrix (T-Matrix) Generation}}, and {\it{Transit-Clustering (TC)}}. These three steps are continuously iterated until an iteration threshold \(N\) is reached or the medoids --- a station representing the center of a cluster --- returned between iterations do not change, outputting the final \(K_1\) clusters, \(C^N_{1,1},C^N_{1,2},... C^N_{1,K_1}\). Our algorithm is an evolution of the ones presented in \cite{Li2015,Li2020}. A detailed discussion of parallelism is presented in Section~\ref{sec:paralellism}. 
The {\bf{AdaTC\(_+\)}} algorithm framework is described in Figure \ref{fig:fig4}.

\subsection{Methodology}
\subsubsection{Geo-Clustering}
Clusters stations into \(K_1\) groups by K-Medoids, based on their geographical distances and check-out patterns. The dissimilarity measure between a pair of stations \(\left(S_h, S_k \right)\) can be seen as a trade-off between the geographical distance and check-out pattern and it is defined as: 

\begin{equation}
\label{eq:eq1}
diss_{GC}(S_h,S_k)= \rho_1 \times gd_{hk} + out_{hk},
\end{equation}
where \(\rho_1 \) is a trade-off parameter,
\(gd_{hk}\) is the geographical distance between \(S_h\) and \(S_k\), and \(out_{hk}\) is the check-out difference between \(S_h\) and \(S_k\), which can be defined as: 

\begin{equation}
\label{eq:eq2}
out_{hk} = || U_h - U_k ||_2 
\end{equation}
where \(U_h\) is a vector with as many entries as the defined time slots that characterize the check-out pattern at a station \(S_h\), {\it{i.e.}},

\begin{equation}
\label{eq:eq3} 
U_h = \left( \frac{u_1}{\sum\limits_{j=1:3}{u_j}},\frac{u_2}{\sum\limits_{j=1:3}{u_j}},\frac{u_3}{\sum\limits_{j=1:3}{u_j}}, \frac{u_4}{\sum\limits_{j=4:5}{u_j}}, \frac{u_5}{\sum\limits_{j=4:5}{u_j}} \right),
\end{equation}
where \(u_i\) is the average check-out at \(S\) in the \(i\)-th time slot (see Table \ref{tab:tab1}). This quantity needs to be estimated and in turn, the estimation of \(u_i\) involves the estimation of another quantity, \(\hat{TL}_{i,t}^S\), the estimated time length with bikes available in station \(S\) in the period \(t\) of the time slot \(i\). This estimation process is explained in detail in Section \ref{sec:sec1}.

In the first iteration, {\it{GC}} is conducted for all stations in the system generating \(K_1\) clusters, \(C^1_{1,1},C^1_{1,2},...,\)
\(C^1_{1,K_1}\). In the following ones, this step is applied on the stations in each cluster obtained in the {\it{TC}} of the previous iteration, {\it{e.g.}}, assuming that the number of stations in each cluster obtained by {\it{TC}} is \(n_1\), \(n_2\) ,..., \(n_{K_2}\), {\it{GC}} clusters the stations in each cluster into \(
[\frac{n_1 \times K_1}{n}],
[\frac{n_2 \times K_1}{n}] ,..., [\frac{n_{K_2} \times K_1}{n}] \)
groups respectively, where \([.]\) is the rounding operator and \(n\) the total number of stations in the system. The number of groups generated in the {\it{GC}} step must be consistent throughout all the {\bf{AdaTC\(_+\)}} algorithm run and therefore always equal to \(K_1\). Consequently, the rounding to be considered, whether by excess or defect, must be the one that ensures that \(\sum\limits^{K_2}_{i=1}\left[\frac{n_i \times K_1}{n}\right] =K_1 \).

{\it{GC}} is executed until convergence {\it{i.e.}}, until a maximum iteration threshold, \textit{max iters gc} (see Figure \ref{fig:fig4}), is reached or the medoids do not change in two consecutive iterations.

\paragraph{Estimate the Average Check-Out at Station \(S\) in the \(i\)-th Time Slot, \(\hat u_i^S\)} \label{sec:sec1}
The historical data of past trips show evidence that the duration of a trip is, on average, much less than the hourly range considered in each time slot (see Table \ref{tab:tab1}), and therefore, each time slot was partitioned into periods of 60 minutes. Since there is not a ground truth regarding the real check-out measure in each station, it was adopted a measure used in previous works \cite{Li2020}, based on the historical trips and bike availability data for each station, to estimate it by: 
\begin{equation}
\label{eq:eq9} 
\hat u_i^S=\frac{1}{\bar{TL}_i^S} \cdot |t| \cdot \bar{r}_i^S,
\end{equation}
where \(\bar{r}_i^S\) is the average number of bikes rented from \(S\) in the time slot \(i\) over all its periods and \(|t|\) the time length of each period considered (\textit{i.e.,} \(|t|= 60 \) minutes). \(\bar{TL}_i^S\) is the average time length, in minutes, in time slot \(i\), over all its periods, that station \(S\) has available bikes to rent.

Let \(r_ {i,t}^S\) be the number of bikes rented from \(S\) in the period \(t\) of the time slot \(i\), \(\hat{TL}_ {i,t}^S\) the estimated time length with bikes available in station \(S\) in the period \(t\) of the time slot \(i\), such that, \( \,  \hat{TL}_ {i,t}^S \leq 60 \: \forall i,t\), and \(np_i\) the number of periods of 60 minutes inside the time slot \(i\). Then, 

 \begin{align}\label{} 
        \bar{r}_i^S = \frac{1}{np_i}\sum\limits_{t=0}^{np_i - 1}{r_ {i,t}^S}
        &&
        \bar{TL}_i^S = \frac{1}{np_i}\sum\limits_{t=0}^{np_i - 1}{\hat{TL}_ {i,t}^S}
\end{align}

\begin{figure}[!t]
\centering
\includegraphics[width=3.1in]{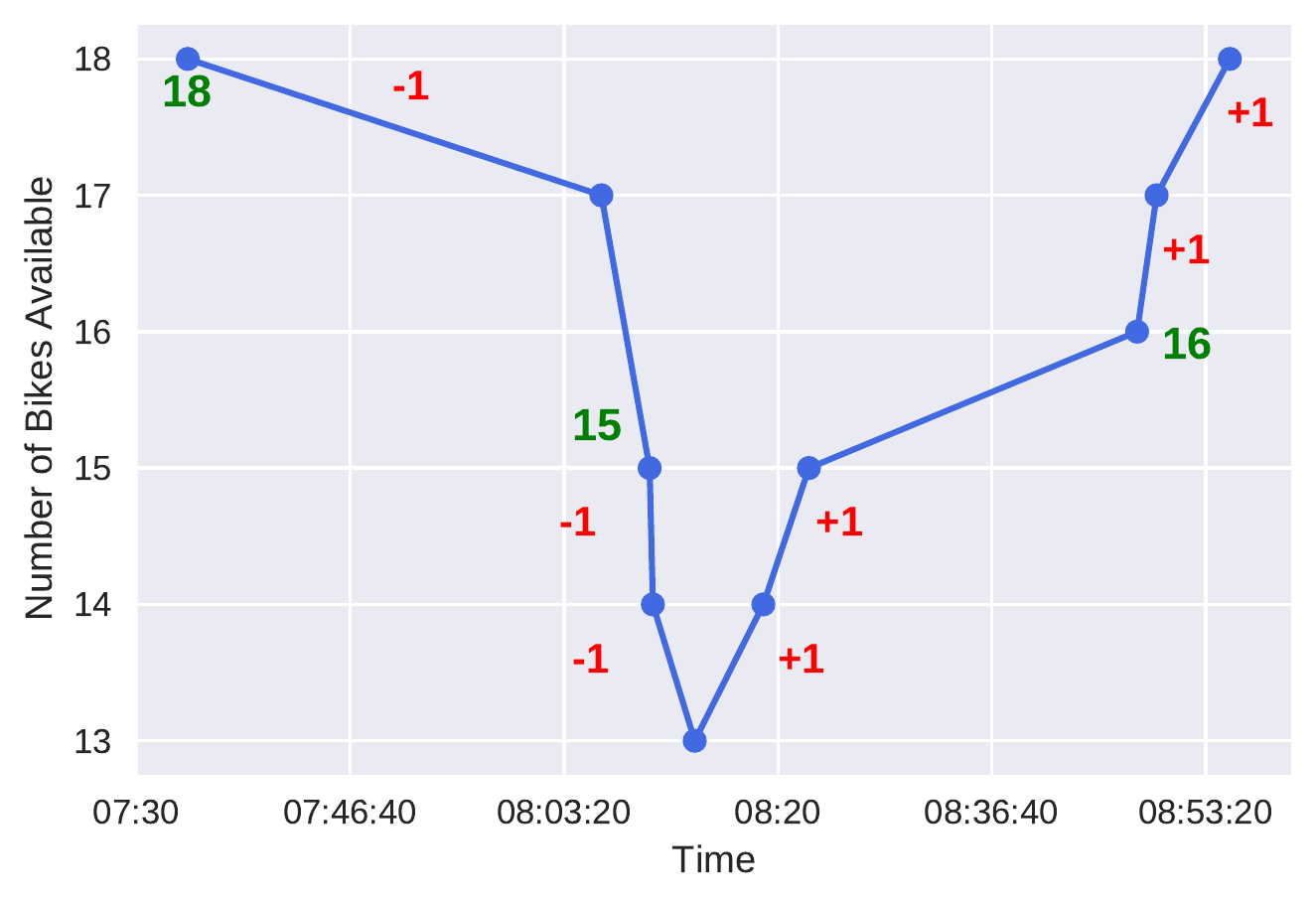}
\caption{Bike Flow in the {\it{W 52 St \& 11 Ave}} Station.}
\label{fig:fig14}
\end{figure}

While \(r_ {i,t}^S\) is directly inferred from the {\it{CitiBike}} trip data, \(\hat{TL}_{i,t}^S\) needs to be estimated.
To estimate \( \hat{TL}_{i,t}^S\) more accurately, besides the {\it{CitiBike}} trip data, it was also taken into account the Open Bus ({\it{OB}}) data \cite{Open:Bus} (see Section \ref{sec:sec2}). Some stations do not have any records on {\it{OB}}, and for these, the estimation of \(\hat{TL}_{i,t}^S\) was performed considering only the {\it{CitiBike}} data. For a given station \(S\), in a period \(t\) of the \(i\)-th time slot, the estimation of \( \hat{TL}_{i,t}^S\) starts by inferring the number of bikes available in each timestamp record resulting from the union of both data sources ({\it{CitiBike}} and {\it{OB}}). This is done for all particular dates, divided according to week and weekend days, in the datasets union with records for the station, time slot, and period under analysis. For the stations with status data on {\it{OB}}, based on the two datasets, it is possible to know exactly the flow of bikes, through the balance (with -1 score for rents and +1 score for returns) and availability at each timestamp. Figure \ref{fig:fig14} exemplifies this procedure for the station {\it{W 52 St \& 11 Ave}} in the time slot [7:00 am - 11:00 am] and the period between 8:00 am and 9:00 am on January 2, 2018.
The {\bf{\textcolor{forestgreen}{green values}}} represent the information available in the {\it{OB}} dataset, while the {\bf{\textcolor{red}{red values}}} the scores for rents and returns according to \textit{CitiBike}. For example, at 7:30 am this station had 18 bikes available (according to {\it{OB}}) and there was a ride that started at this station (according to {\it{CitiBike}}) producing a score of -1 and totaling 17 bikes available at the station after this transition. Following this reasoning, the number of available bicycles was obtained, represented by the {\bf{\textcolor{blue}{blue line}}} in the plot.

For the stations without status data in {\it{OB}}, the number of bikes available in each timestamp was estimated by the {\bf{"Offset" Method}}: summing the absolute value of the minimum (offset) obtained from the cumulative sum, that results from the balance of bikes (again with -1 score for rents and +1 score for returns).

Let \(\hat{TL}_{d,i,t}^S\) be the estimated time length with bikes available in station \(S\), in period \(t\) of \(i\)-th time slot in a specific \(d\) date.
\(\hat{TL}_{d,i,t}^S\) was calculated by adding the length of time between records with the number of bikes available higher than 0.
Then, \( \hat{TL}_{i,t}^S\) is given by:
\begin{equation}
\label{eq:eq10}
{\hat{TL}_{i,t}^S=} \begin{cases}
\frac{\sum\limits_{\text{weekday dates}}{\hat{TL}_{d,i,t}^S}}{\text{weekdays in 2018}},&{\text{for}}\ i=1,2,3\\ 
\frac{\sum\limits_{\text{weekend dates}}{\hat{TL}_{d,i,t}^S}}{\text{weekend days in 2018}},&{\text{for}}\ i=4,5 
\end{cases}
\end{equation}

\subsubsection{Transit-Matrix Generation}
\begin{figure}[!t]
\centering
\includegraphics[width=3.4in]{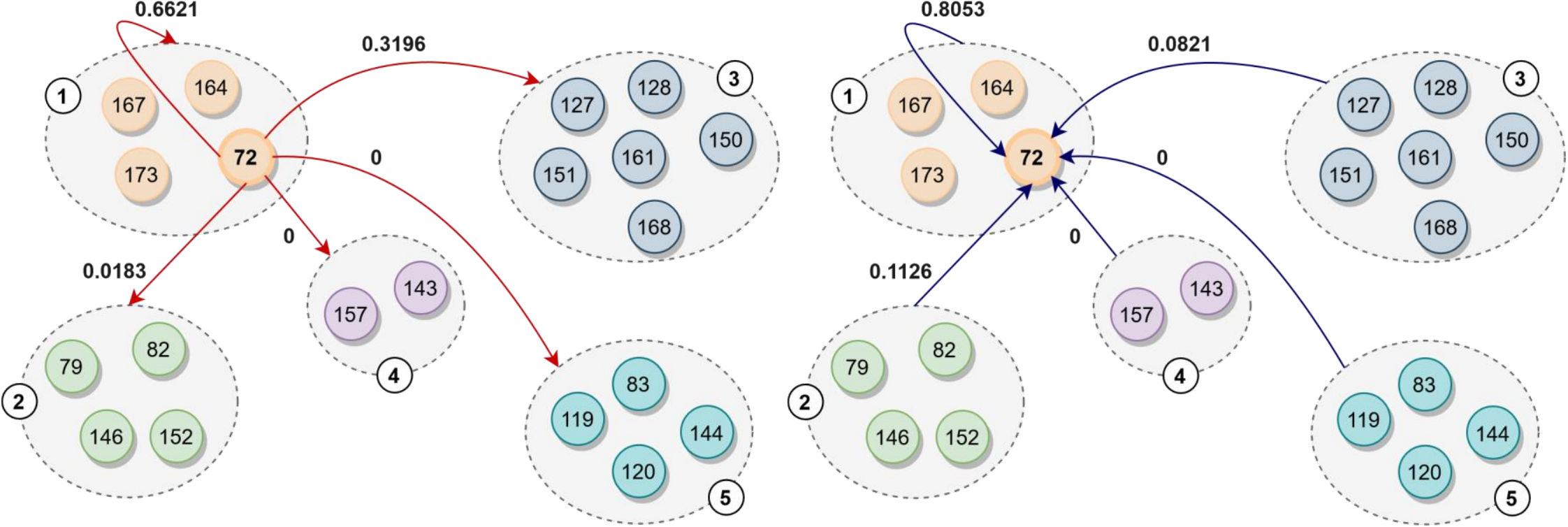}
\caption{{\it{Ride-To-Cluster}} (\emph{{\bf{Left}}}) and {\it{Return-From-Cluster}} (\emph{{\bf{Right}}}) Transitions in \(ts_1\).}
\label{fig:fig5}
\end{figure}

Based on the \(K_1\) clusters obtained by {\it{GC}}, \(n\) {\it{T-Matrices}} are generated, one for each station. Each {\it{T-Matrix}} describes the intra and inter-cluster transitions patterns of a specific station regarding all the time slots. A {\it{T-Matrix}} has 5 rows, each one regarding a time slot presented in Table \ref{tab:tab1}, and each row is a vector with \(2 K_1\) entries, resulting from the concatenation of two other probabilistic \(K_1\)-vectors, {\it{Ride-To-Cluster}} and {\it{Return-From-Cluster}}, in a specific time slot and estimated from the historical trip data. 
{\it{Ride-To-Cluster}} represents how probable is for a bike rented in station \(S\) to be
returned in each of the \(K_1\) clusters, and {\it{Return-From-Cluster}} represents how probable is for a bike returned to station \(S\) to have been rented on each of the \(K_1\) clusters.
For example, for the station with \(id=72\), \(K_1=5\), and considering only 20 stations in the system, Figure \ref{fig:fig5} illustrates the  {\it{Ride-To}} and {\it{Return-From-Cluster}} transitions in the morning rush hours time slot. Therefore, the first row of the corresponding {\it{T-Matrix}} is given by $ (0.662, 0.018, 0.32, 0, 0, 0.81, 0.113, 0.082, 0,0) $.

\subsubsection{Transit-Clustering}
Clusters stations into \(K_2\) groups by K-Medoids, such that \(K_1 \geq K_2\), based on their respective {\it{T-Matrices}} obtained in the previous step. The dissimilarity measure between a pair of stations \(\left(S_h, S_k\right)\) is defined as:

\begin{equation}
\label{eq:eq4}
diss_{TC}(S_h,S_k)= ||A_h - A_k ||_F = ||A_{(h-k)}||_F 
\end{equation}
where \(A_h\) and \(A_k\) are the {\it{T-Matrices}} of stations \(S_h\) and \(S_k\), respectively. \(A_{(h-k)} \) is a matrix with \(5\)$\times$\(2K_{1}\) dimensions that represents the element-by-element subtraction of matrix \(A_h\) with \(A_k\) and whose Frobenius norm is calculated. {\it{TC}} is executed until convergence {\it{i.e.}}, until a maximum iteration threshold, {\it{max iters tc}}, is reached or the medoids do not change in two consecutive iterations.

\subsection{Parallelism with Other Works} \label{sec:paralellism}
The {\bf{AdaTC\(_+\)}} Algorithm implemented in this work was strongly inspired by the Bipartite Clustering (presented in \cite{Li2015}) and AdaTC (presented in \cite{Li2020}) algorithms.
Based on these, we introduced the following modifications:
\begin{itemize}
    \item We considered a trade-off between geographical distance and transition patterns in the {\it{GC}} step dissimilarity function;
    \item The {\it{GC}} and {\it{TC}} steps are based on K-Medoids;
    \item The {\it{T-Matrices}} for each station in the {\it{T-Matrix}} Generation step are constructed differently, differing in their dimensions and entries' meaning;
    \item We derived a new method to estimate the time that a station had bicycles available;
     \item We introduced the convergence parameters and derived a new method to validate both the intrinsic and convergence parameters.
\end{itemize}

%%%%%%%%%%%%%%%%%%%%%%%%%%%%%%%%%%%%%%%%%%%%%%%

\section{Link Prediction Model}

Link Prediction (LP) is a problem of predicting future or missing relationships {\it{(links)}} between entities {\it{(nodes)}} based on network-structured data. In this work is used the {\bf{GraphSAGE}} (SAmple and aggreGatE) algorithm, proposed by Hamilton et al. \cite{Hamilton2017} for building a LP Model, that predicts bike trips based on the learning of node features to consequently create proper node embedding. The problem can be treated as a supervised LP problem.
{\bf{GraphSAGE}} is an {\it{inductive}} node embedding framework \cite{Leskovec}, {\it{i.e.}}, that does not require all graph nodes to be in the network during the training of the embeddings, allowing to generate embeddings for unseen nodes, based on the features and neighborhood of the node, without the need of a re-training procedure. This is extremely useful for dynamic networks that evolve through time, in which new nodes are very likely to integrate the network, allowing to still make predictions, which would not be possible using heuristic methods.
The heuristic methods are a class of simple and effective approaches to LP that computes node similarity scores as the likelihood of links, relative to the network topology and depend on strong assumptions regarding the existence of links. 
By using a GNN it is possible to learn the network features from local subgraphs to create a proper embedding.

\subsection{Calibrate a GraphSAGE LP Model for SBSS}
The 2018 data was used to obtain {\it{four}} clustering configuration scenarios and train {\it{five}} LP models, which in turn is also evaluated on a test hold out data. The 2019 data was considered exclusively to test the LP models predictions for 2019, trained on 2018 and based on the clustering results obtained for 2018, which are assumed not to change significantly between consecutive years. 
Not all the historical trips records registered for 2019 were incorporated in the 2019 test network (see Table \ref{tab:tab5}). In fact, in 2019 there was an increase of 17\% in the number of trips and of 64\% in the number of stations compared to 2018, however, the network for 2019 does not include new stations (nor their trips) that are not in the 2018 data, as clustering cannot be inferred for these (ending up with the configuration described in column \(\mathbf{2019_\text{\bf{restricted}}}\) in Table \ref{tab:tab5}).

\begin{table}[!t]
\caption{Entire Graph Dimensions in 2018 and 2019. \\ \small{\(\mathbf{2019_\text{\bf{restricted}}}\) represents the 2019 graph restricted to the stations (and corresponding trips) in the 2018 setting}. \label{tab:tab5}}
\centering
\resizebox{0.94\linewidth}{!}{%
\begin{tabular}{|cc|cc|cc|}
\hline
 \multicolumn{2}{|c|}{\(\mathbf{2018}\)}                & \multicolumn{2}{c|}{\(\mathbf{2019_\text{\bf{restricted}}}\)}    & \multicolumn{2}{c|}{\(\mathbf{2019}\)}                 \\ \hline
 \multicolumn{1}{|c|}{\# nodes} & \# links & \multicolumn{1}{c|}{\# nodes} & \# links & \multicolumn{1}{c|}{\# nodes} & \# links \\ \hline
 \multicolumn{1}{|c|}{801} & 17.526.058 & \multicolumn{1}{c|}{762} & 18.674.518 & \multicolumn{1}{c|}{1333} & 20.551.697 \\ \hline
\end{tabular}%
}
\end{table}

\subsubsection{Framework}
To calibrate and train this model in {\it{CitiBike}} SBSS 2018 data, the procedure was as follows:
\begin{enumerate}
    \item Create the multigraph network structure;
    \item Sample the train, validation, and test ground truth sets of links (positive and negative). Create the corresponding graphs with the sampled positive links of each set removed: \(G_{train}, G_{val}\), and \(G_{test}\) respectively;
    \item Build a two-layer {\bf{GraphSAGE}} model that:
    \begin{enumerate}
        \item Takes \(\left(S_h,S_k\right)\) node pairs corresponding to possible bike trips and outputs a pair of node embeddings for each node of the pair by concatenating the node feature of this node with the mean of the node features in its neighborhood;
        \item Feed the node embeddings of the candidate link into a Link Classification Layer that applies the inner product to them to construct the edge embedding;
        \item The edge embeddings obtained are passed through a Dense Link Classification Layer, that applies a sigmoid function to the edge embeddings to obtain the predicted probabilities for these candidate links to integrate the network.
    \end{enumerate}
    \item Investigate the calibration of the predicted probabilities ({\it{i.e.}}, if the true frequency of positive labels is approximately equal to its predicted probability) and calibrate the model in the validation set in case of miscalibration, avoiding introducing bias and recurring to an Isotonic Regression Calibrator.
\end{enumerate}

\begin{figure}[!t]
\centering
\includegraphics[width=3.4in]{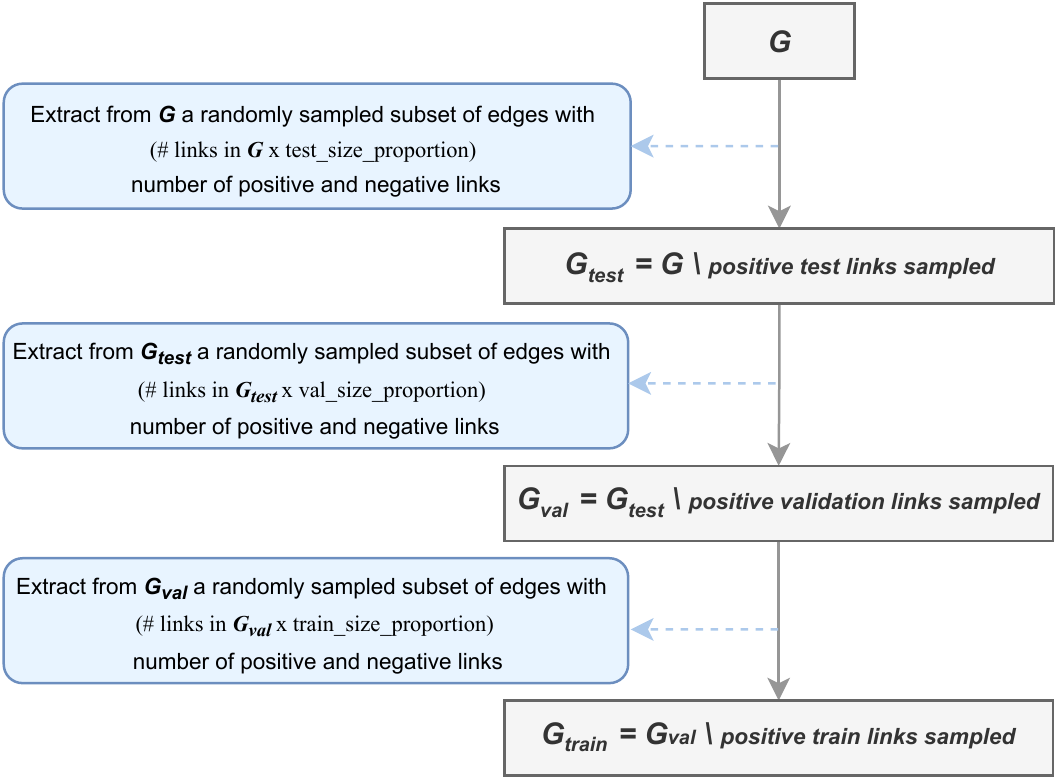}
\caption{Subgraphs Generation and Structure of the Input Graph, \(G\).}
\label{fig:fig13}
\end{figure}

The positive links are those that actually belong to the network and the negative links are links that do not exist in the graph, sampled using Depth-first Search (DFS) at a distance from a source node, selected at random from all nodes in the graph. \(G_{train}\) with the train ground truth set of links, \(G_{test}\) with the test ground truth set of links and \(G_{val}\) with the validation ground truth set of links is used to train, evaluate the performance and calibrate the model, respectively. The split process and subgraphs generation is detailed in Figure \ref{fig:fig13}.
The entire model is trained by minimizing the binary cross-entropy loss between the predicted link probabilities and the true link labels.

\subsubsection{Network Structure}
Let \(G=(V,E)\) be a homogeneous transitions network with nodes representing the stations (with an associated vector of features) and links corresponding to the inter-stations transitions. 
The network structure implemented has all stations in the analysis as nodes with the correspondent trips as links.

Two network structures configurations were tested, {\it{one}} including the clustering results in the node features and the {\it{other}} without considering clustering results as node features. 
The inclusion of the clustering results in the node features could streamline the task of predicting the traffic in each station, given that the cluster information introduces a measure of similarity between the stations and those in the respective neighborhood. 

For the first structure configuration, each node has a feature vector with 34 entries, consisting of:
\begin{itemize}
    \item Group ID of the cluster to which the station belongs according to a given clustering algorithm;
    \item Average Air Temperature ($^{\circ}$F), Relative Humidity (\%), Wind Speed (mph), One-hour precipitation (mm), and Visibility (mi) in the four seasons: Spring, Summer, Autumn, and Winter; 
    \item Relative frequency histogram of trips going out the node during the days of the week and weekend (from Monday to Sunday);
    \item Relative frequency histogram of trips going out the node in six predefined time periods: 
    \newline
    [7:00am - 11:00am[, [11:00am - 3:00 pm[, [3:00 pm - 7:00 pm[, [7:00 pm - 11:00 pm[, [11:00 pm - 3:00 am[, [3:00 am - 7:00 am[.
\end{itemize}
In the second structure type, each node has the same features, except for the Group ID of the cluster.

\section{Evaluation Experiments}
\subsection{Data Collection and Baselines}\label{sec:sec2}

\subsubsection{Datasets}

\paragraph{Historical Trips}
{\it{CitiBike}} makes available the system data \cite{CB} of its users' historical trips, containing information such as trip duration, trip' start and end timestamp, official {\it{CitiBike}} ID, name, latitude, and longitude for the start and end station, etc.

\paragraph{Meteorology}
Reports of airport weather conditions, crucial for aviation, are a source of historical weather data. The weather data was collected from the Iowa Environmental Mesonet (IEM) website \cite{IEM}, which collects environmental data synced from the real-time ingest every 10 minutes. 
The network considered for extracting this data was the \emph{New York ASOS Network}, under the New York Time Zone (EST/EDT).

\paragraph{Station Status}
Open Bus \cite{Open:Bus,OB:Guide} collects monthly continuous monitoring records related to the NYC public transportation and their conditions/status, such as the number of available bikes in specific timestamps.

\paragraph{Geographical Distances}
The pairwise distances between all stations were extracted from Open Route Service Maps API \cite{ORS:Maps}, measured in meters and according to the regular cycling profile. 

\begin{figure}[!t]
\centering
\includegraphics[width=2.1in]{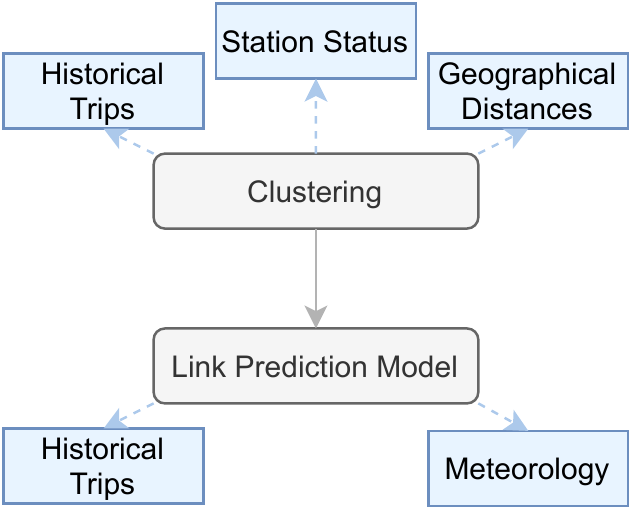}
\caption{The Datasets into the Proposed Framework.}
\label{fig:fig12}
\end{figure}

Figure \ref{fig:fig12} shows the datasets inputted to each part of our two-step framework.
Some of the code implemented and data used in our work is available at \url{https://github.com/barbara20901/cluster-link-prediction/blob/main/README.md}

\subsubsection{Baselines and Metrics}
The method proposed in this work to predict the bike traffic is to implement a LP Problem based on GNN embeddings and built-in a StellarGraph Network \cite{StellarGraph}, which integrates among other indicators, the results of the {\bf{AdaTC\(_+\)}} algorithm. 
To confirm the effectiveness of our proposed methodology, experiments were conducted to compare it with four other baselines, three of them consisting in formulating the same LP problem but integrating the clustering results of the following clustering techniques:
\begin{itemize}
\item {\it{K-Medoids ({\bf{KM})}}} with \(70\) clusters formations and based on \(D_{sym}\), the symmetric dissimilarity matrix between pairs of stations according to the dissimilarity function \(diss_{GC}\) (see Equation \eqref{eq:eq1}) and setting \(\rho_1=0.505\);
\item {\it{Spectral Clustering ({\bf{SC})}}} with \(70\) clusters formations and based on the affinity/similarity matrix \(S = \frac{(1-D_{sym})}{ \textrm{max} \, D_{sym}}\);
\item {\it{Geo-Clustering ({\bf{GC})}}} with \(K_1=70\), \(\rho_1=0.505\) and executed until convergence (first step in {\bf{AdaTC\(_+\)}}).
\end{itemize}
The fourth baseline {\it{No Clustering ({\bf{NC})}}} consists in formulating the same LP problem but without any clustering of stations in the node features.

The metric adopted to measure the baseline results ({\bf{KM, SC, GC}}, and {\bf{NC}}) compared with the proposed methodology ({\bf{AdaTC\(_+\)}}) is the accuracy obtained in each one, after ascertaining a possible model miscalibration. The most accurate model allows us to infer the best clustering structure for the data and the cluster pertinence.

\subsection{AdaTC\(_+\) Clustering}

Before conducting clustering, all stations with a residual percentage of trips (chosen assuming that each station should have at least about half the number of days in one year, in terms of the number of annual trips) compared with the total number of trips in 2018 were removed from the clustering analysis to allow an accurate check-out pattern estimate.
Moreover, the time was discretized into five time slots, according to Table \ref{tab:tab1}, crucial to streamline the implementation of the algorithm, when certain measures need to be calculated that ended up to be aggregated, becoming characterizers of each station. 

\begin{table}[!t]
\caption{Defined Time Slots.\label{tab:tab1}}
\centering
\resizebox{0.85\linewidth}{!}{%
\begin{tabular}{|c|cc|c}
\cline{1-3}
\textbf{Day}             & \textbf{Time}       & \textbf{Time Slot} &     \\ \cline{1-3}
\multirow{3}{*}{Weekday} & 7:00 am - 11:00 am  & Morning Rush Hours \\ \cline{2-3}
                         & 12:00 pm - 4:00 pm & Day Hours  \\ \cline{2-3}
                         & 5:00 pm - 10:00 pm & Evening Rush Hours \\ \cline{1-3}
\multirow{2}{*}{Weekend} & 9:00 am - 5:00 pm  & Trips Hour \\ \cline{2-3}
                         & 6:00 pm - 11:00 pm & Evening Hours  \\ \cline{1-3}
\end{tabular}%
}
\end{table}

\subsubsection{Validating Parameters}

The {\bf{AdaTC\(_+\)}} algorithm has three intrinsic parameters, two of them referring to the {\it{GC}} step and the last to the {\it{TC}} step: \(\rho_1\) and \(K_1\), and \(K_2\), respectively. The ones related to its convergence are \(N\), {\it{max iters gc}}, and {\it{max iters tc}}. 
The convergence parameters were validated by evaluating the behavior of the Total Dissimilarity Function (TDF) over a large number of iterations. 
For {\it{max iters gc}} and {\it{max iters tc}}, the TDF is the sum of the dissimilarity between each data point (\(n\) stations) and the medoid of the cluster to which they were assigned to (\(m_{i,k}\)), according to the dissimilarity function defined in the respective step, {\it{i.e.}},
\begin{equation}
\label{eq:eq12}
TD_{GC}=\sum\limits_{i=1}^{n}{diss_{GC}(S_i,m_{i,k})}
\end{equation}

\begin{equation}
\label{eq:eq13}
TD_{TC}=\sum\limits_{i=1}^{n}{diss_{TC}(S_i,m_{i,k})}
\end{equation}

For \(N\), the TDF is the sum of the TDF for {\it{GC}}, \(TD_{GC}\), with the TDF for {\it{TC}}, \(TD_{TC}\). For sake of space, we do not show here our experiments in the convergence parameters validation.

The {\bf{AdaTC\(_+\)}} algorithm was conducted with the convergence parameters validated for several different combinations of the intrinsic parameters. The combinations were generated by setting \(\rho_1\) to vary on a linear scale and their midpoints, with:
\begin{align*}
\label{eq:eq6}
\rho_1 & \in  \{0,5\times10^{-4},10^{-3},5.5\times10^{-3},10^{-2}, \\ & 5.5\times10^{-2},10^{-1},5.05\times10^{-1},1,5.5,10\} \\
K_1  & \in \{50,60,70,80,90,100\} \\
K_2  & \in  \{10,20,30,40,50\}. 
\end{align*}

The results of 330 combinations were evaluated according to the following four metrics:
\begin{enumerate}
    \item {\bf{Between all pairs of stations in the same cluster}}
        \begin{itemize}
        \item Average Inner-Cluster Geographical Distance ($\mathbf{AGD}_\textrm{inner}$)
        \item Average Inner-Cluster Check-Out Difference ($\mathbf{ACOD}_\textrm{inner}$)
        \end{itemize}
    \item {\bf{Between all the medoids of each cluster}}
        \begin{itemize}
        \item Average Inter-Cluster Geographical Distance ($\mathbf{AGD}_\textrm{inter}$)
        \item Average Inter-Cluster Check-Out Difference ($\mathbf{ACOD}_\textrm{inter}$)
        \end{itemize}
\end{enumerate}

\begin{figure}[!t]
\centering
\includegraphics[width=3.4in]{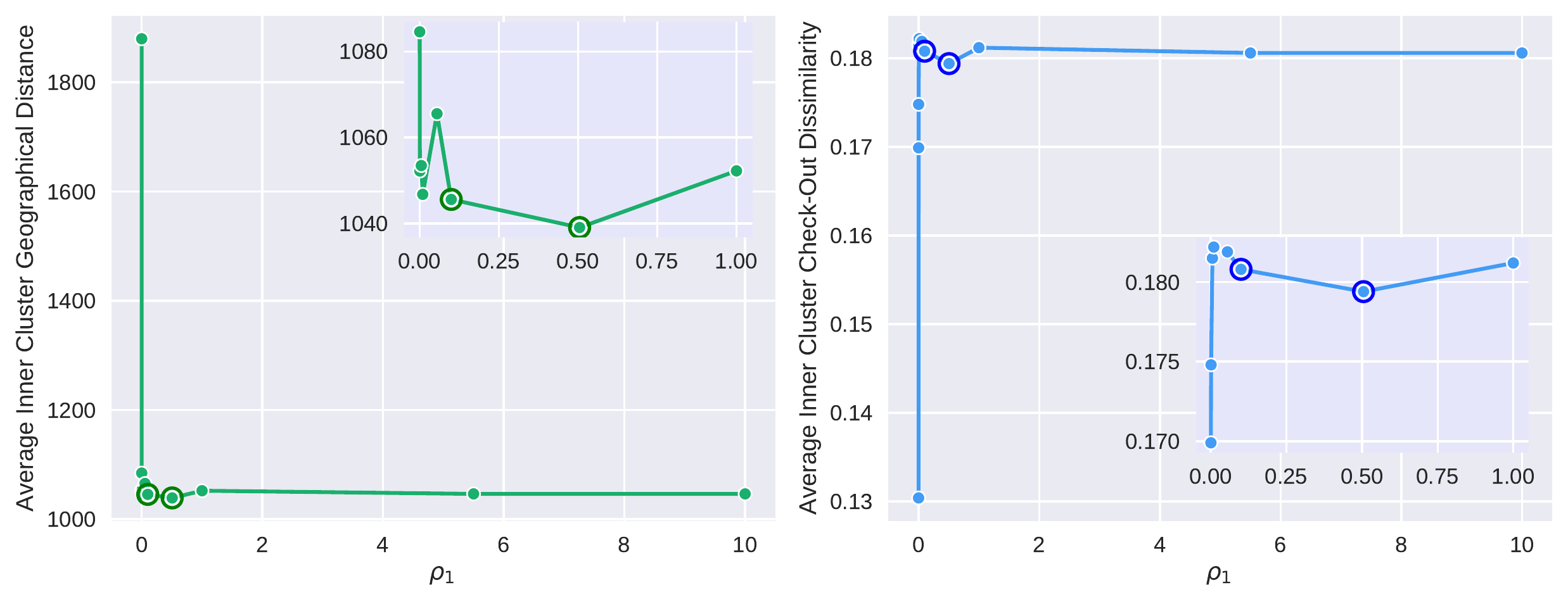}
\caption{\(\mathbf{AGD_\textrm{inner}}\) and \(\mathbf{ACOD_\textrm{inner}}\) for the \(\rho_1\) tested values. The values for  \(\rho_1=0.1\) and \(\rho_1=0.505\) are highlighted.}
\label{fig:fig6}
\end{figure}

Figure \ref{fig:fig6} shows the behavior of the average {\it{inner-cluster}} metrics mentioned for all \(\rho_1\) tested values averaged over \(K_1\) and \(K_2\), and can be seen as a marginal distribution in one dimension. Its analysis provides some intuition apriori to the value of the \(\rho_1\). An increasing \(\rho_1\) makes the \(\mathbf{AGD_\textrm{inner}}\) decrease while the \(\mathbf{ACOD_\textrm{inner}}\) increase, which is quite reasonable considering its trade-off role. This happens because, for a fixed \(K_1\) and \(K_2\), a lower \(\rho_1\) value does not prioritize geographical distance, and as such, clusters are not geographically as visible and organized, causing the average {\it{inner distance}} to increase and the {\it{inner check-out dissimilarity}} to decrease.

With this idea in mind, it is important to choose a \(\rho_1\) value that minimizes both the {\it{inner measures}}, since it is intended to partition the area under study in small functional regions with similar transition patterns, and accordingly, the most appropriate \(\rho_1\) value seems to be 0.1 or 0.505, given that they are an elbow point on the {\it{inner distance}} plot and have a corresponding reasonable {\it{inner check-out dissimilarity}} (being even the ones with the lowest {\it{inner check-out dissimilarity}} without considering the smaller \(\rho_1\)). Analyzing the metrics results as a function of the three intrinsic parameters (\(\rho_1\), \(K_1\), and \(K_2\)) the best parameters combination was found by restricting the search space to \(\rho_1 \in [0.1; 0.505]\), and analyzing the volume of the space of the parameters such that the combination found presents the best possible performance overall, {\it{i.e.,}} the combination that best minimizes the {\it{inner measures}} and maximizes the {\it{inter measures}} simultaneously. 

The three best solutions found in the grid of the search space considered, ordered by optimality were:
\begin{equation} 
\label{eq:eq7} \\
\begin{cases} 
    \rho_1=0.505  \\
    K_1=90 \\
    K_2=10 \\
\end{cases}
\begin{cases}
    \rho_1=0.505\\
    K_1=70 \\
    K_2=40 \\ 
\end{cases}
\begin{cases}
    \rho_1=0.1\\
    K_1=90 \\
    K_2=10 \\
\end{cases}
\end{equation}

Although the first parameters combination in \eqref{eq:eq7} provides the best solution, the second parameters combination was chosen for having an intermediate \(K_1\) value within the range of the \(K_1\) values tested. It is expected that a larger value of \(K_1\) ({\it{i.e.,}} more clusters with fewer stations within each) may partition the geographical space too much that the similarities between stations are not significant. In turn, a small value of \(K_1\) ({\it{i.e.,}} fewer clusters with more stations within each) may also compromise the bike usage regularity in each cluster.

\subsubsection{Clustering Results}

In our clustering experiments, 801 {\it{CitiBike}} stations were clustered into 70 clusters by {\it{four}} clustering methods: {\bf{AdaTC\(_+\)}}, {\it{Spectral Clustering {\bf{(SC)}}}}, {\it{Geo-Clustering {\bf{(GC)}}}}, and {\it{K-Medoids {\bf{(KM)}}}}.

\begin{figure}[!t]
\centering
\subfloat[]{\includegraphics[width=1.6in]{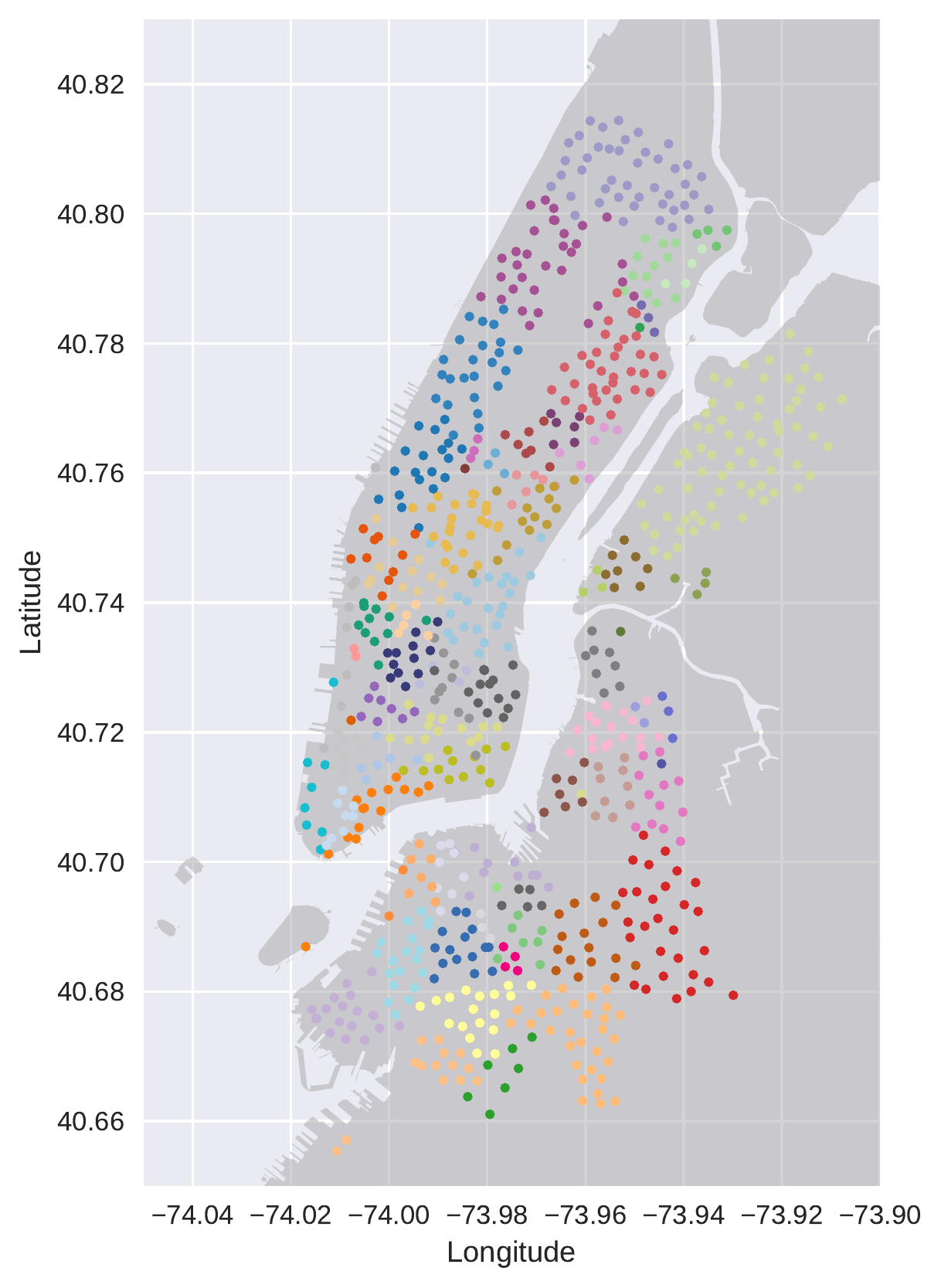}%
\label{fig:fig71}}
\hfil
\subfloat[]{\includegraphics[width=1.6in]{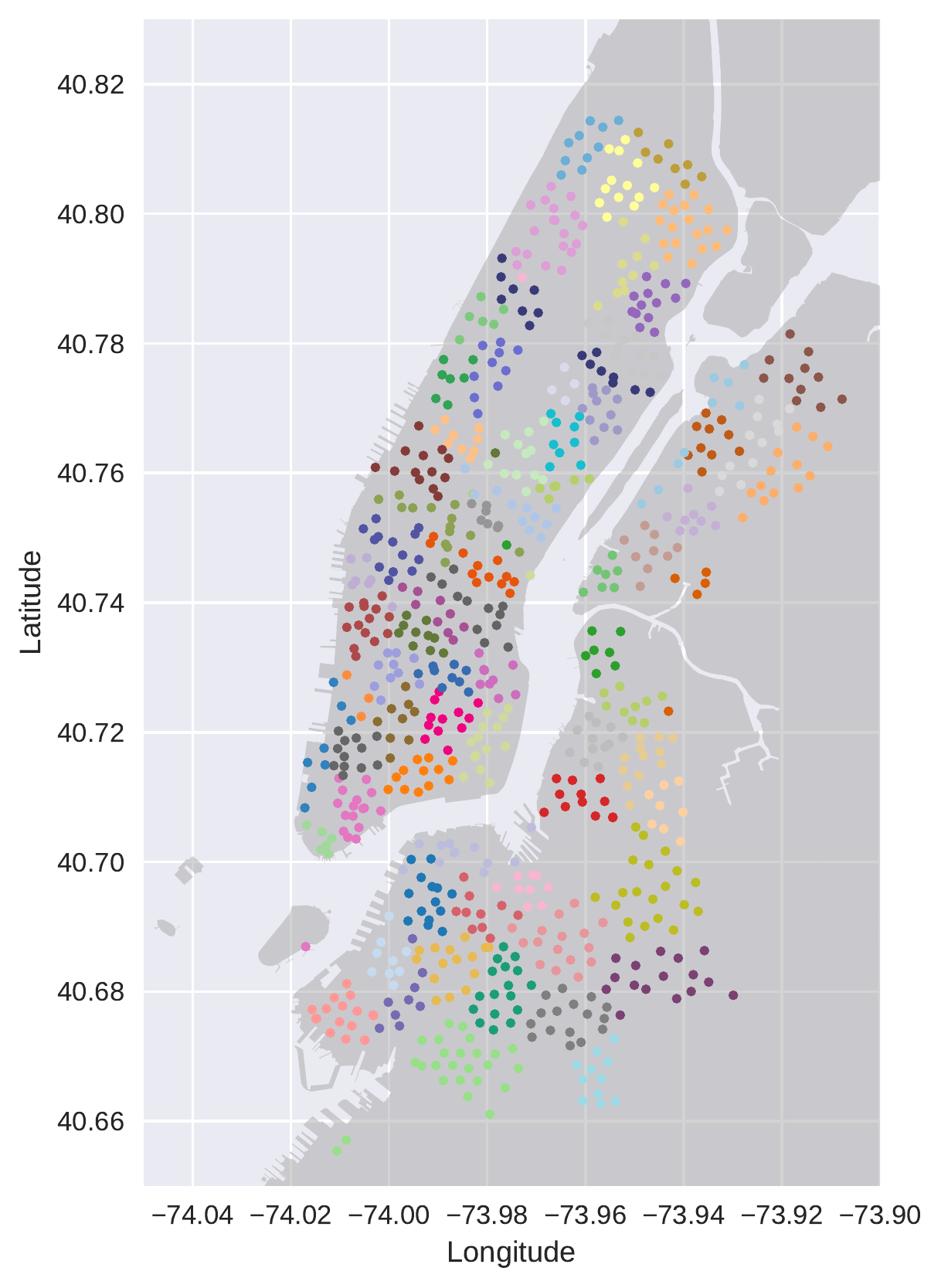}%
\label{fig:fig72}}
\caption{Clustering Configurations Obtained for {\it{CitiBike}} Stations for (a) {\bf{AdaTC\(_+\)}} and (b) {\bf{SC}}.}
\label{fig:fig7}
\end{figure}

\begin{figure}[!t]
\centering
\subfloat[]{\includegraphics[width=1.6in]{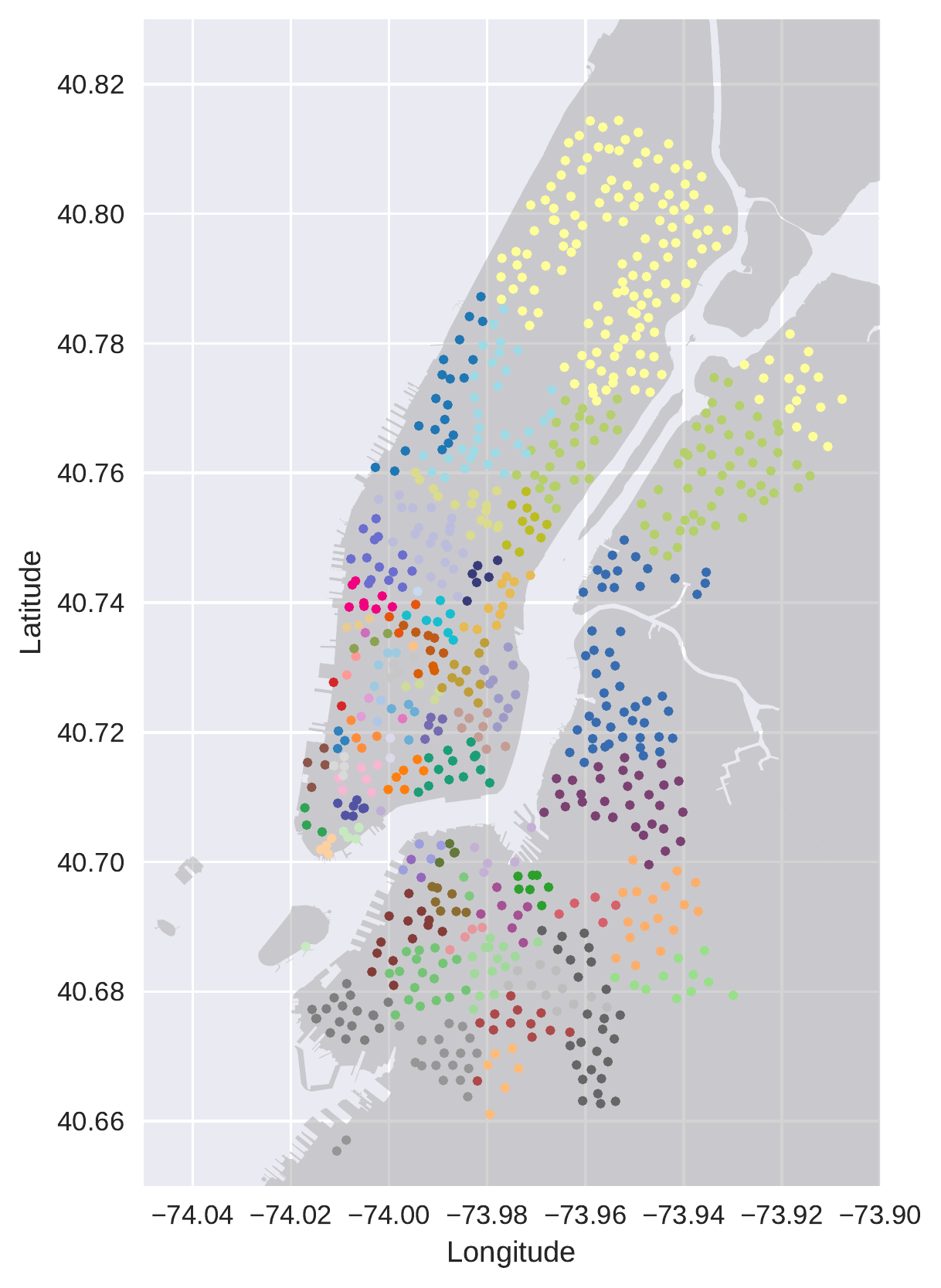}%
\label{fig:fig81}}
\hfil
\subfloat[]{\includegraphics[width=1.6in]{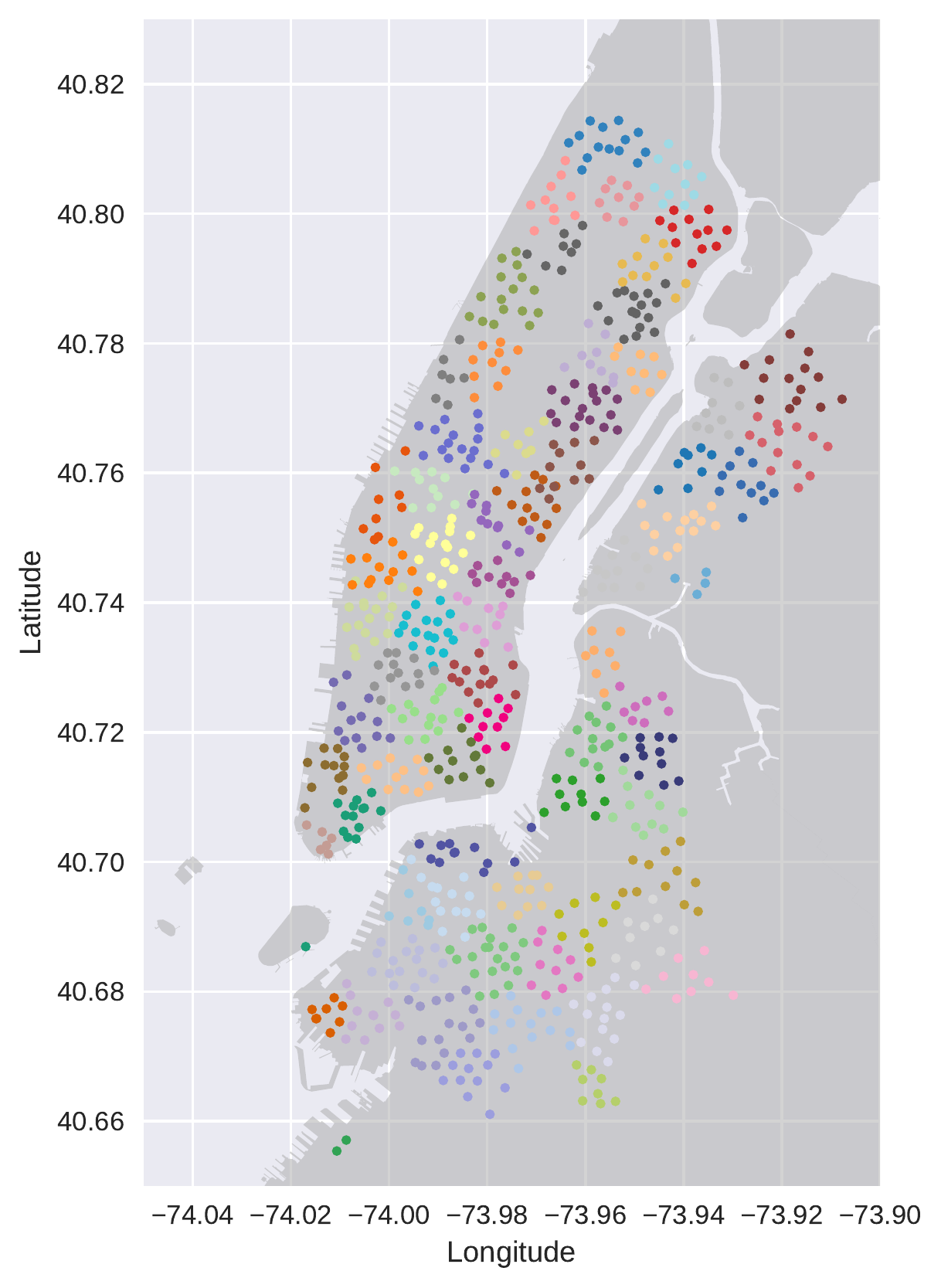}%
\label{fig:fig82}}
\caption{Clustering Configurations Obtained for {\it{CitiBike}} Stations for (a) {\bf{GC}} and (b) {\bf{KM}}.}
\label{fig:fig8}
\end{figure}

The following figures show the results of each clustering method on the {\it{CitiBike}} stations. A brief comment is made on the results of each clustering configuration, but these are merely descriptive since the best model is chosen based on the performance on the prediction task. Figures \ref{fig:fig72} and \ref{fig:fig82} shows that the {\bf{SC}} and {\bf{KM}} algorithms seem to partition the city geographically similarly. Besides, {\bf{GC}} seems unable to detect groups in \textit{North Manhattan}, something the remaining three clustering algorithms have identified (see Figure ~\ref{fig:fig81}). As illustrated in Figure \ref{fig:fig71}, {\bf{AdaTC\(_+\)}} produces clusters geographically more condensed and organized. Furthermore, the respective medoids are meaningful since this algorithm considers as medoids some stations (or its neighboring stations) in the popularity top. This may be due to the {\it{TC}} extra step in the {\bf{AdaTC\(_+\)}} algorithm, which can group geographically close stations into a single cluster by recognizing their identical transition matrices.
It is possible to observe this phenomenon comparing, for example, the further north clusters (\textit{upper Manhattan}) in Figure \ref{fig:fig71} with Figures \ref{fig:fig72} and \ref{fig:fig81}.
Besides, {\bf{AdaTC\(_+\)}} is the one that performs the best on the prediction task, as can be seen later.

\subsection{Link Prediction Results}

In Table \ref{tab:tab3} we present the accuracy obtained for each model in the respective test set for both years under study. The accuracy calculates how often predictions match the binary labels of each link.
Regarding the 2018 test data, despite the small difference, {\bf{AdaTC\(_+\)}} outperforms the baselines. We can confirm our initial hypothesis that the cluster induces interesting and important properties for this problem, by verifying that, for 2018, all baselines with clustering have better results (or equal in the case of {\bf{KM}}) when compared to the {\bf{NC}} baseline. With the test done for the 2019 trips data, we were able to get an idea of how well the trained model generalizes for the following year, even if with only some node features updated. Besides, the results are promising given that the accuracy of the 2019 results has not dropped considerably, even using a 2018 setting, with {\bf{GC}} outperforming (not significantly again) {\bf{AdaTC\(_+\)}} when generalizing for 2019. 

\begin{table}[!t]
\caption{Accuracy in the Test Set after Calibration. AdaTC\(_+\) outperforms the baselines and when in mismatch, the performance does not degrade significantly. \label{tab:tab3}}
\centering
\resizebox{0.85\linewidth}{!}{%
\begin{tabular}{c|c|c|c|c|c|}
\cline{2-6}
 &
    {\bf{AdaTC\(_+\)}} & {\bf{GC}} & {\bf{SC}} & {\bf{KM}} & {\bf{NC}} \\ \hline
\multicolumn{1}{|c|}{\bf{2018}} & {\bf{\textcolor{blue}{88\%}}}          & 87\%        & 86\%        & 83\%        & 83\%        \\ \hline
\multicolumn{1}{|c|}{\bf{2019 (Mismatch)}} & 85\%           & {\bf{\textcolor{blue}{86\%}}}        & 85\%        & 83\%        & 84\%        \\ \hline
\end{tabular}%
}
\end{table}

\begin{figure}[!t]
\centering
\includegraphics[width=3.4in]{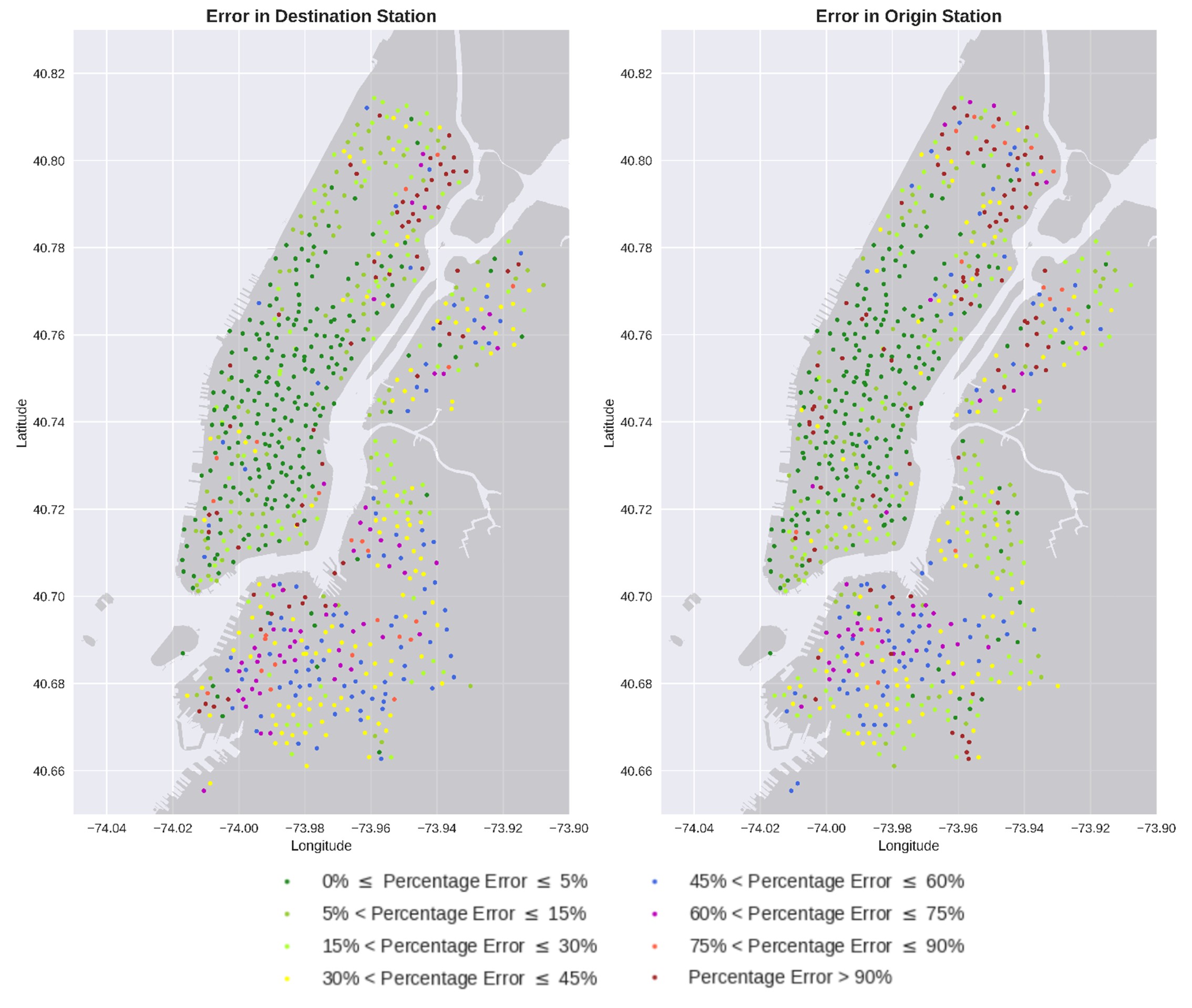}
\caption{Stations colored according to the respective Prediction Error (\%) made by our predictor in 2018. \small{The stations with infinite error i.e., the ground truth number of trips starting (ending) there are null, but the model predicts some, are not represented in this plot.}}
\label{fig:fig9}
\end{figure}

The following figures present a more detailed analysis of the predictions made by our predictor based on {\bf{AdaTC\(_+\)}} for the 2018 test set. Figure \ref{fig:fig9} shows the percentage of the prediction error in relation to the number of ground truth trips in the 2018 test set that start and end in each station vs the predicted.
Let \(x_t\) be the ground truth, {\it{i.e.}}, the number of true rides, and \(x_p\) the number of positive rides predicted by the predictor. The Prediction Error Percentage was calculated for each station at origin and destination according to:

\begin{equation}
\label{eq:eq8}
\mathbf{PE}=\frac{1}{x_t} \left| x_t - x_p\right| \cdot 100
\end{equation}

By the inspection of Figure \ref{fig:fig9}, we can see that our model predicts well the vast majority of the number of trips that start and end in a specific station, namely in \textit{Manhattan}. The reddish colors (more evidenced in the \textit{Brooklyn} and \textit{Queens} zones) are associated with higher prediction error and Table \ref{tab:tab4} helps to understand this behavior. From this Table, it is possible to see that the reddish stations can be seen as outliers stations, given that they are associated with a low number of true trips (starting and ending) in the test set and therefore larger prediction error associated.

\begin{table}[!t]
\caption{Prediction Error Intervals Analysis for Origin and Destination.\label{tab:tab4}}
\centering
\resizebox{0.9\linewidth}{!}{%
\begin{tabular}{|cc|cc|cc|}
\cline{3-6}
\multicolumn{2}{c|}{\multirow{2}{*}{}}             & \multicolumn{2}{c|}{\bf{Origin}} & \multicolumn{2}{c|}{\bf{Destination}} \\ \cline{3-6} 
\multicolumn{2}{c|}{} &
  \multicolumn{1}{c|}{\bf{\# Stations}} &
  {\bf{\begin{tabular}[c]{@{}c@{}}Mean Value \\ of Real Trips\end{tabular}}} &
  \multicolumn{1}{c|}{\bf{\# Stations}} &
  {\bf{\begin{tabular}[c]{@{}c@{}}Mean Value \\ of Real Trips\end{tabular}}} \\ \hline
\multicolumn{1}{|c|}{\multirow{8}{*}{\bf{\begin{tabular}[c]{@{}c@{}}Prediction \\ Error\\  Intervals\end{tabular}}}} &
  \textbf{{[}0\%-5\%{]}} &
  \multicolumn{1}{c|}{206} &
  152 &
  \multicolumn{1}{c|}{223} &
  133 \\ \cline{2-6} 
\multicolumn{1}{|c|}{} & {\bf{{]}5\%-15\%{]}}}    & \multicolumn{1}{c|}{110}   & 63.25   & \multicolumn{1}{c|}{98}       & 65        \\ \cline{2-6} 
\multicolumn{1}{|c|}{} & {\bf{{]}15\%-30\%{]}}}   & \multicolumn{1}{c|}{129}   & 38.3    & \multicolumn{1}{c|}{108}      & 48.87     \\ \cline{2-6} 
\multicolumn{1}{|c|}{} & {\bf{{]}30\%-45\%{]}}}   & \multicolumn{1}{c|}{105}   & 33.87   & \multicolumn{1}{c|}{116}      & 31.8      \\ \cline{2-6} 
\multicolumn{1}{|c|}{} & {\bf{{{]}45\%-60\%{]}}}}   & \multicolumn{1}{c|}{80}    & 33.7    & \multicolumn{1}{c|}{89}       & 35.26     \\ \cline{2-6} 
\multicolumn{1}{|c|}{} & {\bf{{]}60\%-75\%{]}}}   & \multicolumn{1}{c|}{42}    & 50.9    & \multicolumn{1}{c|}{54}       & 45.57     \\ \cline{2-6} 
\multicolumn{1}{|c|}{} & {\bf{{]}75\%-90\%{]}}}   & \multicolumn{1}{c|}{16}    & 27.5    & \multicolumn{1}{c|}{23}       & 28.4      \\ \cline{2-6} 
\multicolumn{1}{|c|}{} & {\bf{{\textgreater 90\%}}} & \multicolumn{1}{c|}{82}    & 5.73    & \multicolumn{1}{c|}{59}       & 10.15     \\ \hline
\end{tabular}%
}
\end{table}

\begin{figure}[!t]
\centering %start
\includegraphics[width=3in]{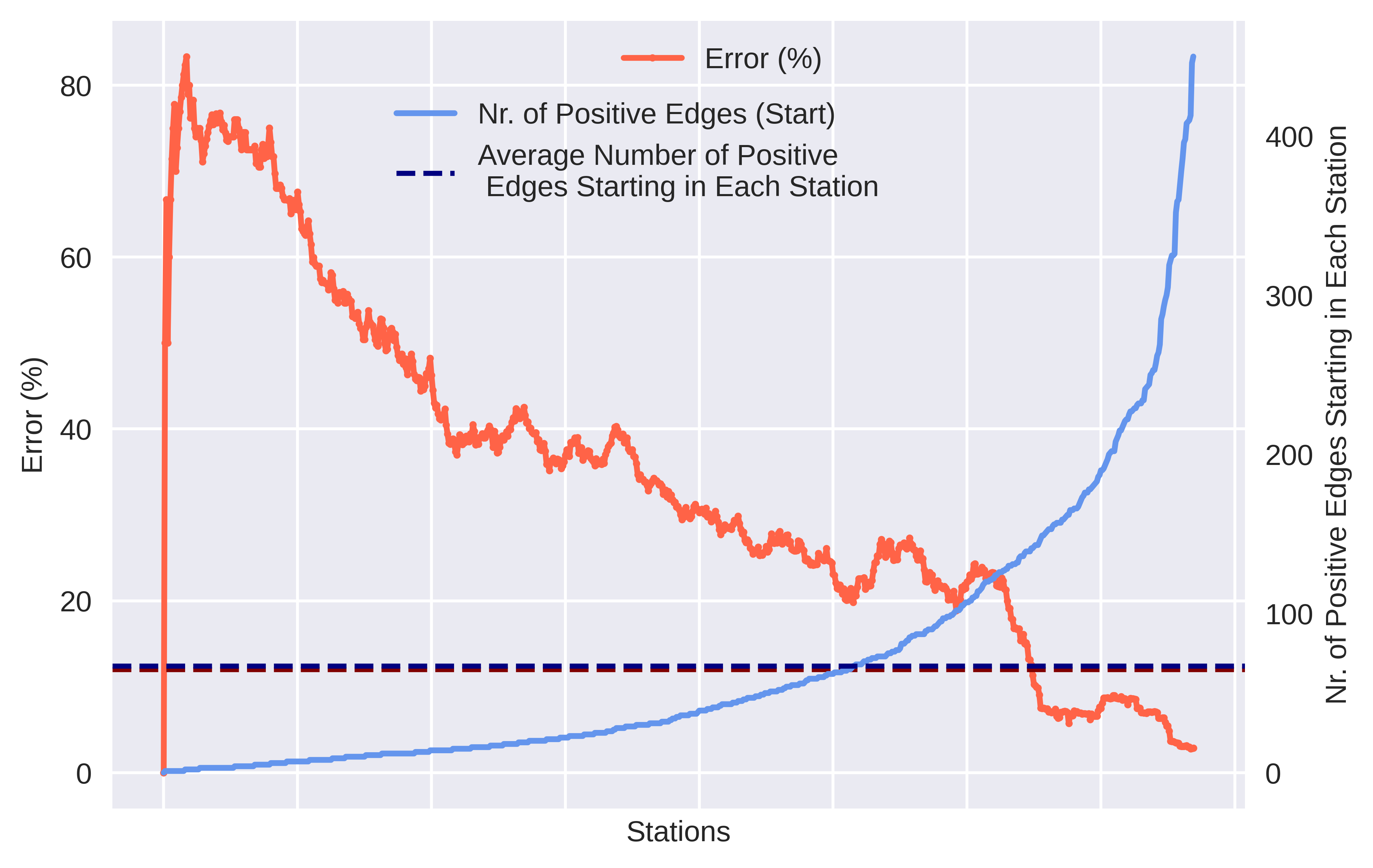}
\caption{Error and Edges in Origin Stations. The blue and red dashed horizontal lines represent the number of 67 trips and the 12\% of error, respectively.}
\label{fig:fig10}
\end{figure}

\begin{figure}[!t]
\centering %stop
\includegraphics[width=3in]{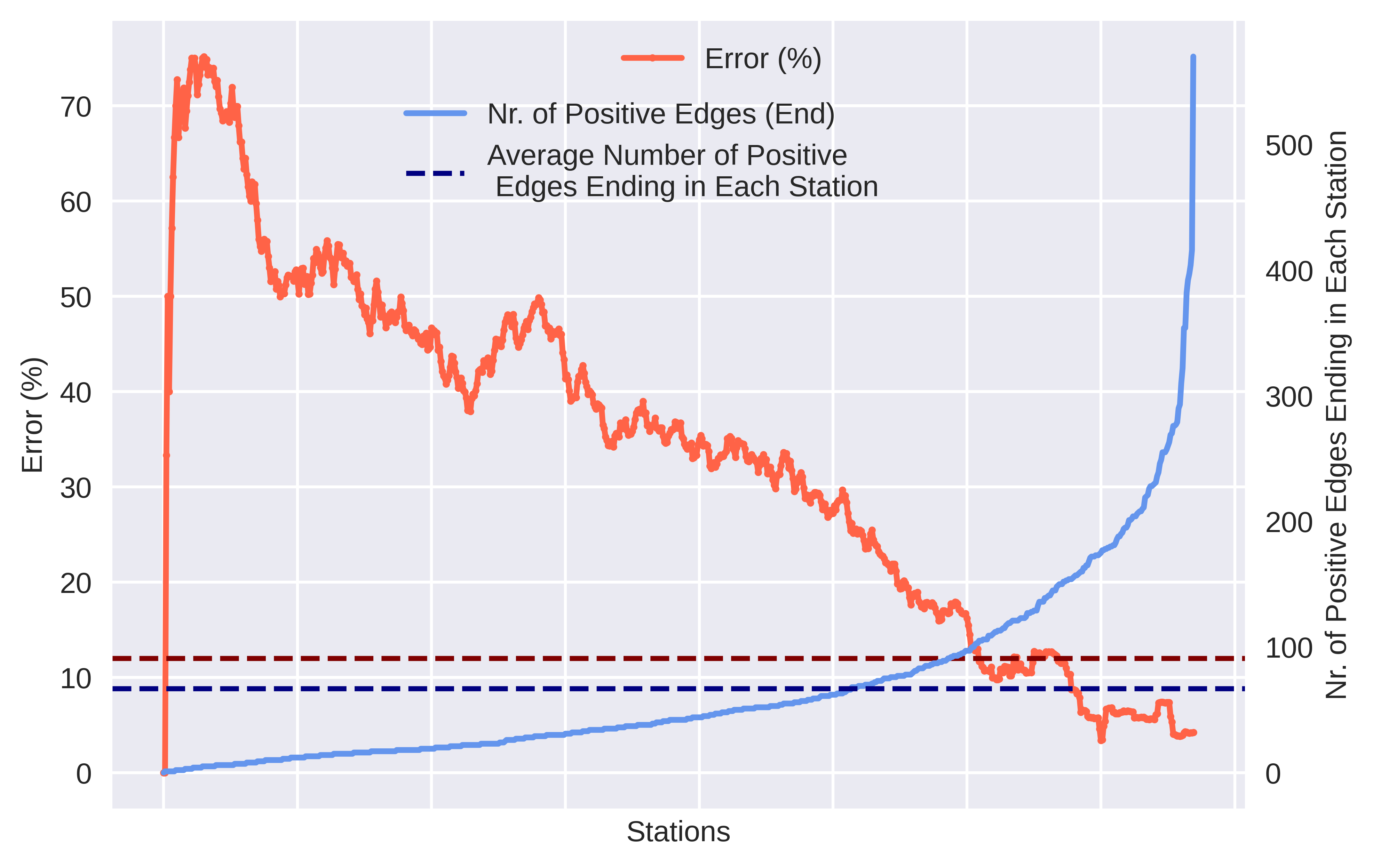}
\caption{Error and Edges in Destination Stations. The blue and red dashed horizontal lines represent the number of 67 trips and the 12\% of error, respectively.}
\label{fig:fig11}
\end{figure}

To further investigate the nature of the higher errors in the test set, we plotted the prediction error percentage for each station at the origin and destination (Figures \ref{fig:fig10} and \ref{fig:fig11}, respectively), and to be possible to interpret this percentage, the stations in the x-axis were sorted by the increasing number of trips, so we can clearly see the downward trend of error as the number of trips increase. 
By analyzing Figures \ref{fig:fig10} and \ref{fig:fig11}, we can ensure that for stations with stable transition patterns, {\it{i.e.,}} at least more than 67 trips starting and ending in that station (the average of the number of trips starting and ending in the test set for all stations), the prediction error for origin and destination stations does not exceed around 20\%, which on average represents 20\% of error. The average of the number of trips starting and ending in the test set for all stations can be seen as an order of magnitude from which the model works well, {\it{i.e.,}}, for stations with a number of trips above the average, the order of magnitude of the error is in accordance with the order of magnitude of the error of our predictor.
Besides, the 12\% of error in the predictor for 2018 (according to Table \ref{tab:tab3}) is verified for stations with the number of starting trips higher or equal to around 120 and the number of ending trips higher or equal to around 100 and can be seen as an upper bound on the error in this scenario. Another interesting aspect rises from Figures \ref{fig:fig10} and \ref{fig:fig11}. Once again, we have evidenced the imbalanced problem we are facing. The intersection between the blue lines indicates that, on average, 64\% of the stations have the number of trips starting and ending below the average value in the test set (514 of the 801 stations under analysis). These stations with atypical patterns may be influencing the accuracy when in reality it is intended to ensure that there is supply at the most chaotic stations (where demand can overcome supply). Therefore, the model can only improve if we further restrict the stations to be considered.

\section{Discussion and Conclusions}
Our work addresses the traffic prediction problem in a SBSS. We provide a framework on how to predict future transition patterns, based on past information. These predictions can be the basis of rebalancing strategies, as they reflect the flow of bicycles through the area under study with a given precision, and clustering reveals the inter and intra-clusters dynamics, which is crucial and meaningful for both \textbf{System Prediction} and \textbf{Operation} branches.
Our proposed LP model based on {\bf{AdaTC\(_+\)}} combined with {\bf{GraphSAGE}} outperformed the remaining, maintaining stability even under harsh mismatch. We confirmed our initial hypothesis that clustering induces interesting and important properties including the robustness that a GNN and its embeddings leverage to address this problem. Our approach has a main limitation: for atypical years, in which the system settings change significantly ({\it{e.g.}}, many new stations), we cannot take advantage of the {\it{inductive}} nature of the model as in this situation it will not be possible to infer the new stations' cluster. In this case, there is no other solution than to remove the new stations from the analysis and integrate them into the training of the following year. As future work, it would be interesting to develop a strategy of {\it{Continual Learning}} to overcome this limitation.
Besides the network configurations mentioned, another network configuration has been tested. This configuration consisted of considering as nodes only the medoids of each cluster and the corresponding links. The smaller size of this network forced to greatly reduce the proportion of edges to be sampled, and the training of the model ended up with a poor accuracy of 56\%. If this approach had succeeded, it would be a way to significantly reduce the problem complexity (in terms of the number of stations).

We have empirically shown that the combination of {\bf{AdaTC\(_+\)}} and a GNN-based Link Predictor can achieve stable and accurate predictions of bike trips in a SBSS, from weather, distance, and historical trip data. 

\bibliographystyle{IEEEtran}
\bibliography{ref.bib}

\section*{Biography Section}
\begin{IEEEbiographynophoto}{Bárbara Tavares} is a M.Sc. student in Data Science and Engineering at Instituto Superior Técnico. Her B.Sc. degree is in Applied Mathematics in the Statistics and Operation Research Branch at the Faculty of Sciences of the University of Lisbon. She considers herself a data science enthusiast and is fascinated by solving real-world problems.
\end{IEEEbiographynophoto}

\begin{IEEEbiographynophoto}{Cláudia Soares} holds a Ph.D., M.Sc. and B.Sc. in Electrical and Computer Engineering from Instituto Superior Técnico, Portugal, and a degree in modern languages and literature, from Nova University of Lisbon. Prof. Soares is an Assistant Professor at NOVA School of Science and Technology, and a researcher at NOVA LINCS, Portugal. She uses real-world data problems to identify the shortcomings of current machine learning, data science, and Big Data methods. She applies optimization, statistics, and probability theory to address those gaps, developing robust, interpretable, and fair learning methods that can be trusted in real life. Her application areas are in healthcare, transportation, environmental and urban sciences, and space. 
\end{IEEEbiographynophoto}

\begin{IEEEbiographynophoto}{Manuel Marques} is a Researcher in the Electrical and Computer Engineering Department at Instituto Superior Técnico (IST) and also at the Instituto de Sistemas e Robótica (ISR). He holds a Ph.D. in Electrical and Computer Engineering from IST (2011) and his main area of research is computer vision with special interest on 3D reconstruction, object recognition and video processing. Recently, Dr. Marques focus on applications to assistive robotics and transportation, in particular, cycling mobility.
\end{IEEEbiographynophoto}
%biography text

\vfill

\end{document}